\documentclass[11pt]{article}

\usepackage[utf8]{inputenc}
\usepackage[T1]{fontenc}
\usepackage[a4paper,margin=1in]{geometry}
\usepackage{graphicx}
\usepackage{amsmath,amssymb}
\usepackage{booktabs}
\usepackage{caption}
\usepackage{float}
\usepackage{hyperref}
\usepackage{microtype}
\usepackage[scale=0.96]{tgheros}
\usepackage{titling}
\usepackage[super,sort&compress]{natbib}

\setcitestyle{super,sort&compress,citesep={,}}

\graphicspath{{./}}
\hypersetup{
  colorlinks=true,
  linkcolor=black,
  citecolor=black,
  urlcolor=blue
}

\microtypesetup{expansion=false}
\pretitle{\begin{center}\fontsize{22}{24}\selectfont\bfseries}
\posttitle{\par\end{center}\vspace{-0.8em}}
\preauthor{\begin{center}\normalsize}
\postauthor{\par\end{center}\vspace{-1.2em}}

\title{Culturally uneven urban perception in large language models}
\author{Rong Zhao$^{1,2,\dagger,*}$, Wanqi Liu$^{1,\dagger}$, Zhizhou Sha$^{3}$, Nanxi Su$^{2}$, Yecheng Zhang$^{2,*}$, Ying Long$^{2}$\\[0.5em]
\small $^{1}$Centre for Advanced Spatial Analysis (CASA), UCL, London, UK\\
\small $^{2}$School of Architecture, Tsinghua University, Beijing, China\\
\small $^{3}$Department of Computer Science, UT Austin, Austin, TX, USA\\[0.5em]
\small $^{\dagger}$These authors contributed equally to this work.\\
\small $^{*}$Correspondence to: rong.zhao.25@ucl.ac.uk; zhangyec23@mails.tsinghua.edu.cn}
\date{}

\begin{document}

\maketitle
\begin{abstract}
Large language models (LLMs) are increasingly used to describe and evaluate cities, yet the cultural structure of their urban judgments remains understudied. Here we introduce a measurement framework for testing whether LLM-based urban perception is culturally neutral, using a globally stratified street-view image dataset. Open-ended descriptions and structured scores generated by three frontier multimodal models all show that the neutral baseline lies closer to regional framings associated with Europe and North America than to other cultural framings. Comparisons between AI and human urban perception further show that prompting can move AI responses closer to specific regional human descriptions, but fails to recover the variety and diversity of human responses, flattening observed demographic patterns and introducing sentiment-based self-favouring bias. These results indicate a systematic risk in treating AI as a neutral tool for urban tasks, especially when model outputs are used to compare, evaluate or represent cities across cultural contexts.
\end{abstract}

\section*{Introduction}

Urban perception is a core way of understanding how people interpret, evaluate and attach meaning to the built environment. From \textit{The Image of the City} to later work on place and the evaluative image of urban scenes, city perception has been treated not only as a response to physical form, but also as a judgment about the social and evaluative meaning of place\cite{lynch1960image,relph1976place,nasar1990evaluative}. This idea has since been operationalised at scale. Crowdsourced image-comparison studies showed that perceived safety and related qualities could be mapped across cities, turning subjective urban perception into a measurable object of computational urban analysis\cite{salesses2013collaborative}.

This large-scale measurement tradition has depended on human perception data. Deep-learning approaches extended pairwise urban-image perception datasets by predicting scores from street-view imagery, allowing safety, beauty, wealth and other perceived qualities to be estimated across much larger urban regions\cite{dubey2016deep}. Subsequent work applied similar methods to large-scale urban environments and consolidated visual urban perception as a growing field of urban analytics\cite{zhang2018measuring,ito2024understanding,globalstreetscapes2024}. Yet these methods still require human judgments as their empirical foundation, whether through crowdsourced comparisons, survey responses or labelled perception datasets. The rise of large language models and multimodal models changes this measurement problem. Instead of only training models on human perception labels, researchers can now ask AI systems to directly describe, score or explain urban scenes. Recent studies have used generative or vision-language models to evaluate place identity, urban attractiveness, safety, streetscape change and other visible qualities of urban environments, while broader tests suggest that GenAI models can capture urban-science patterns but may oversimplify urban complexity\cite{jang2024place,malekzadeh2025urban,beneduce2025urban,zhang2025genai,xiao2025chatgpt4o}. Other work has begun to benchmark or calibrate multimodal models for street-view-based urban perception, streetscape assessment, cycling environments, human-preference alignment and broader urban intelligence tasks\cite{mushkani2025benchmark,zhang2026urbanalign,morenovera2025urbanvlm,perez2025sagai,han2026streetperception,wu2026cycling,zhang2025urban_safety,huang2024zeroshot,feng2025urbanllava}.

The convenience of AI-based urban perception does not remove the cultural structure of perception itself. Classic cross-cultural studies show that visual preferences for urban street scenes differ across cultural settings, rather than reflecting a single universal aesthetic standard\cite{nasar1984visual}. Environmental preference also varies across cultural and sub-cultural groups\cite{kaplan1987cultural}. More recent large-scale evidence makes this limitation more consequential: global urban visual perception varies across demographics, personalities and geographic contexts, rather than forming a single universal standard of safety, beauty, liveliness or desirability\cite{quintana2025global}. Related work further shows that image-based urban-perception models can be plausible but misleading when transferred to Chinese urban landscapes without careful adaptation\cite{rui2025plausible}. If human perception is culturally and demographically heterogeneous, then any AI system used to represent ``urban perception'' must be examined for whose perception it approximates and whose it suppresses.

This concern is reinforced by evidence that large language models are themselves culturally and socially uneven. LLMs can approximate some human judgments and cognitive regularities, which makes them tempting tools for simulating human responses\cite{binz2025foundation,steyvers2025knowledge}. However, they also encode geographic and cultural biases, align unevenly with cultural values and reproduce social-group distortions\cite{manvi2024geobias,tao2024culturalalignment,lu2025cultural,hofmann2024dialect}. When used as substitutes for human participants, LLMs can misportray and flatten identity groups, reducing heterogeneous social perspectives into simplified model-generated patterns\cite{wang2025flatten}. Related work has raised parallel concerns about cultural erasure, social identity bias and the difficulty of measuring culture in model outputs\cite{qadri2025erasure,hu2025socialidentity,adilazuarda2024culture,alkhamissi2024culturalalignment,li2024culturellm}. These findings echo the long-standing critique that behavioural research has often treated Western, educated, industrialized, rich and democratic populations as an implicit human default\cite{henrich2010weirdest}, and support the view that large AI models are cultural and social technologies rather than neutral prediction systems\cite{farrell2025cultural}. Prompting a model to adopt a social or cultural role does not resolve this issue, because role play in LLMs should not be equated with genuine standpoint-taking\cite{shanahan2023roleplay}.

These literatures create a specific tension for AI-based urban perception. On one side, LLMs and multimodal models appear to offer a scalable alternative to conventional human-centred perception measurement. They can generate open-ended descriptions and structured scores for large numbers of urban scenes, making them attractive for urban research, planning and design\cite{fu2025llmplanning,batty2025generative}. On the other side, both human urban perception and LLM outputs are culturally structured. Existing studies of LLM-based urban perception have mostly asked whether models can score, benchmark or calibrate urban scenes against human preferences\cite{mushkani2025benchmark,zhang2026urbanalign,morenovera2025urbanvlm}. Existing studies of LLM cultural bias have mostly relied on text-based prompts, cultural surveys or social scenarios detached from concrete urban environments\cite{tao2024culturalalignment,morehouse2025bias}. What remains unclear is whether LLM-based urban perception is culturally neutral when models interpret actual urban scenes: what the model treats as a general view of the city, how regional cultural framing shifts that view, and whether these shifts recover or suppress the heterogeneity observed in human perception.

Here we introduce a measurement framework for testing cultural non-neutrality in LLM-based urban perception. We use a globally stratified street-view image dataset as a controlled visual test bed, allowing the same urban scenes to be evaluated repeatedly under different model and prompt conditions. We compare three frontier multimodal models under a neutral prompt and seven regional cultural framings. This design makes the model's apparently neutral response empirically comparable with responses generated under explicit cultural framings, allowing us to test whether the neutral condition behaves as a culturally placeless reference point or as a culturally uneven one.

The framework combines two complementary task formats. In the open-ended task, models describe the same urban scenes in free text. We measure semantic distance from the neutral response, examine the geometry of prompt-conditioned descriptions, estimate sentiment-based ingroup preference and compare model descriptions with human text-image descriptions from Geograph Britain and Ireland. In the structured task, models score each scene on six dimensions: safety, liveliness, wealth, beauty, boredom and depression. We then compare these scores with external human-perception baselines and with qscore-based demographic subgroup differences from an external human-perception dataset. Together, these comparisons test not only whether cultural framing changes model outputs, but also whether such changes bring model perception closer to human perception or instead compress human variation.

Across both task formats, we show that LLM-based urban perception is culturally uneven. The neutral baseline lies closer to regional framings associated with Europe and North America than to other cultural framings. Regional prompts can move model outputs toward specific human reference descriptions, but they do not recover the diversity of human responses. Model descriptions remain more uniform and more positive than human descriptions, and structured model scores often flatten demographic patterns observed in human perception. These findings indicate a systematic risk in treating AI as a neutral tool for urban perception, especially when model outputs are used to compare, evaluate or represent cities across cultural contexts.

\section*{Results}

\begin{figure}[H]
  \centering
  \includegraphics[width=\textwidth,height=0.82\textheight,keepaspectratio]{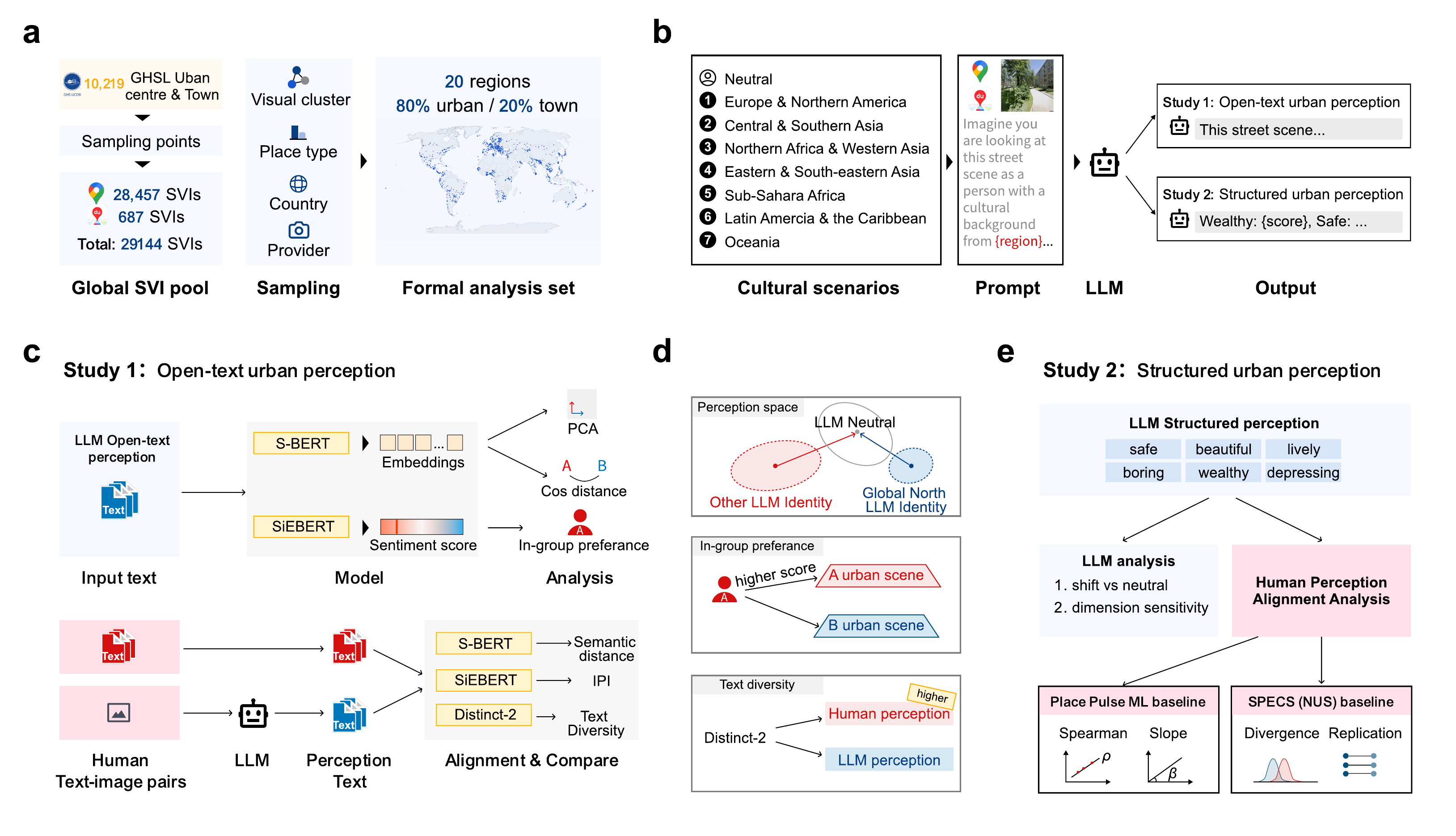}
  \caption{\textbf{Study design for testing culturally uneven LLM-based urban perception.} a, Construction of the global street-view image (SVI) pool and stratified sampling into a 3,000-scene formal analysis set. b, Neutral and regional cultural prompting used for both open-text and structured perception tasks. c, Open-text analysis pipeline, including semantic embedding, local geometry, sentiment scoring, ingroup preference and human text-image comparison. d, Conceptual summaries of the main open-text outputs: semantic position relative to neutrality, sentiment-based ingroup preference and text-diversity comparison. e, Structured six-dimension perception task and external grounding against human-perception baselines.}
  \label{fig:fig1}
\end{figure}

The analysis begins with the sampling pipeline in Fig.~\ref{fig:fig1}a. We combined GHSL urban-centre and town locations with Google and Baidu street-view imagery, producing a 29,144-image pool (28,457 Google SVIs and 687 Baidu SVIs) with explicit provider provenance and audit metadata. From this pool, we drew a 3,000-image formal analysis set, stratifying by visual cluster, place type, country and provider so that comparisons across 20 regions did not collapse onto a small set of recurring streetscape forms.

The prompting design then split the same image set into two complementary tasks (Fig.~\ref{fig:fig1}b). The seven Meso7 cultural framings are Europe and Northern America (ENA), Central and Southern Asia (CSA), Northern Africa and Western Asia (NAWA), Eastern and South-eastern Asia (ESEA), Sub-Saharan Africa (SSA), Latin America and the Caribbean (LAC), and Oceania (OCE); these abbreviations are used hereafter. In the open-text task, three LLMs described each scene under a neutral prompt and these seven regional contexts, yielding 72,000 descriptions for the semantic-distance, local-geometry and sentiment-based ingroup-preference analyses shown schematically in Fig.~\ref{fig:fig1}c. Those model descriptions were then compared with human text-image pairs from the Geograph Britain and Ireland project to ask whether culturally proximate prompting improved alignment with human place descriptions or instead exposed a different form of model compression (Fig.~\ref{fig:fig1}d). In parallel, the structured task asked the same models to score each scene on six explicit dimensions--safe, lively, wealthy, beautiful, boring and depressing--creating 72,000 evaluations for prompt-shift analysis, external grounding and qscore-based comparison with publicly released human-perception data (Fig.~\ref{fig:fig1}e).

\subsection*{Study 1 | Open-ended urban perception}

\begin{figure}[p]
  \centering
  \includegraphics[width=\textwidth,height=0.78\textheight,keepaspectratio]{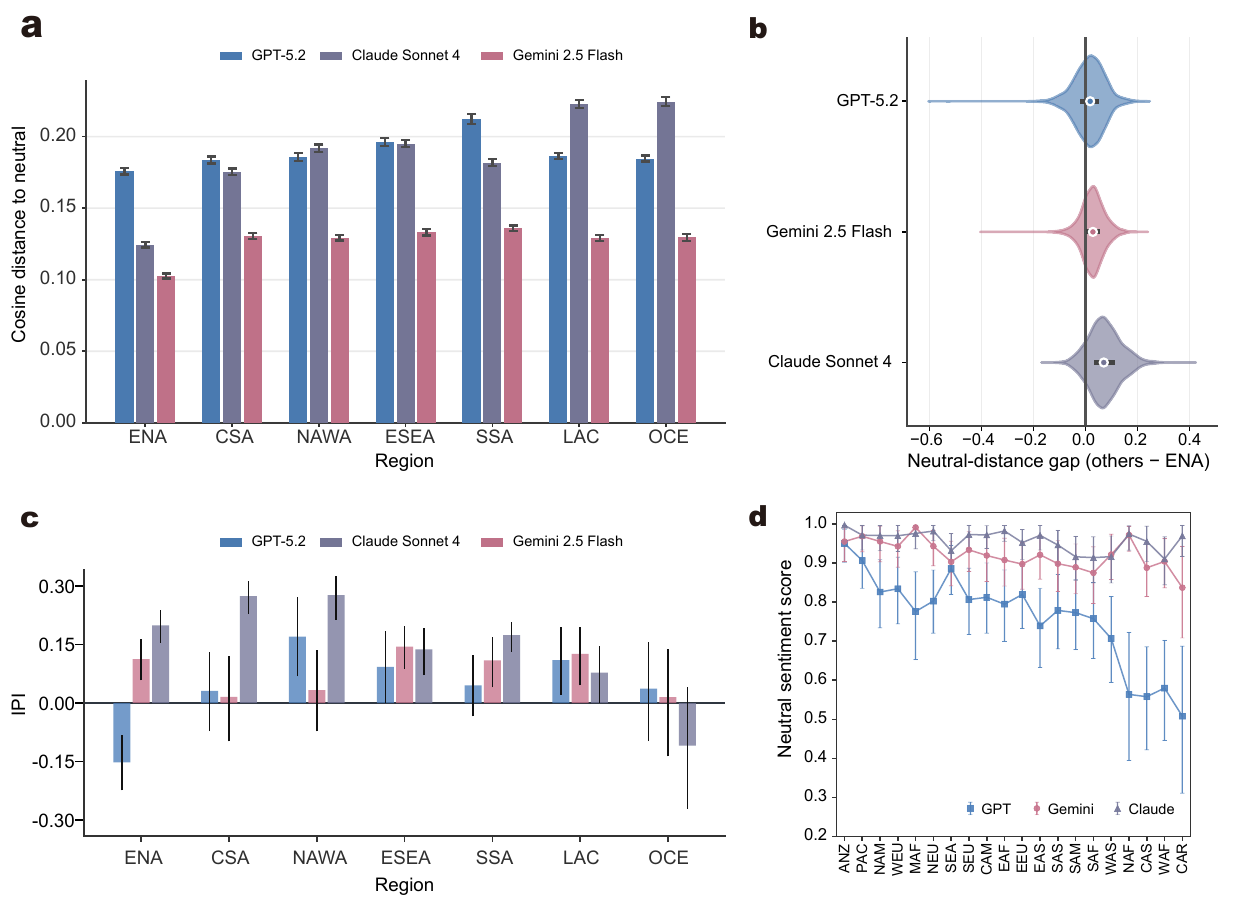}
  \caption{\textbf{Open-ended city perception shows a culturally uneven neutral baseline and region-linked affective evaluation.} a, Mean semantic distance between each regional prompt and the matched neutral response. b, Bootstrap distributions of the neutral-distance gap between ENA and the other regional prompts. c, Region-level sentiment-based ingroup preference index (IPI) estimates with 95\% bootstrap confidence intervals for each model. d, Neutral-prompt sentiment scores across 20 scene micro-regions, ordered from higher to lower mean sentiment across the three models. Regional abbreviations are defined in Supplementary Table~S5.}
  \label{fig:fig2}
\end{figure}

We begin with open-ended scene descriptions, which allow the models to decide which visual and social cues matter. For each image, we compared the description generated under each regional cultural framing with the description generated under the neutral prompt. Across all three models, the neutral prompt did not behave as a culturally invariant semantic baseline (Fig.~\ref{fig:fig2}a,b). ENA was consistently closest to the neutral response, with a model-mean cosine distance of 0.134. The other regional framings averaged 0.174, making ENA about 23\% closer to the baseline. LAC and OCE were the most displaced conditions, both averaging 0.180. The same ordering held across all models, although the overall spread varied across systems.

This pattern means that the baseline is not simply a midpoint among cultural framings. The bootstrap gap distributions in Fig.~\ref{fig:fig2}b compare each regional distance with the ENA distance. They are concentrated above zero, showing that the neutral response lies systematically nearer to some regional lenses than to others. In semantic terms, the models' default description of a street scene is therefore closer to an ENA framing than to many non-Western framings, rather than equally distant from all of them. The same semantic asymmetry appeared when the regional framing was made coarser or finer: ENA was again the closest Macro5 framing, and Northern America and Western Europe were the two closest Micro20 framings by cross-model mean distance (Supplementary Fig.~S3). It also remained visible when the Meso7 analysis was repeated with additional open-weight vision-language models, indicating that the pattern is not limited to the three frontier systems used in the main figure (Supplementary Fig.~S7). A local PCA of the same description embeddings provides a complementary view: regional centroids separate in different directions relative to the neutral condition in each model, indicating that cultural framing changes the orientation of semantic movement as well as its distance from the baseline (Supplementary Fig.~S2).

Affective evaluation showed a related but distinct pattern. The seven-region estimates in Fig.~\ref{fig:fig2}c report a sentiment-based ingroup preference index (IPI): whether a regional framing describes scenes from its own region more positively than other framings describe the same scenes. This self--other tendency was broadly positive: 19 of the 21 model-region estimates were above zero, and six of seven cross-model regional means were positive. The magnitude varied, with the largest regional means in NAWA (0.160), ESEA (0.125), SSA (0.109), CSA (0.107), and LAC (0.105), while OCE was near zero. Coarser Macro5 and finer Micro20 prompt families produced the same broad pattern of positive self--other affective shifts, and the additional-model check again showed that positive IPIs were not confined to a single frontier model family (Supplementary Figs.~S4 and S7).

Sentiment also varied by the regions to which the street scenes belonged (Fig.~\ref{fig:fig2}d). The ordering broadly placed Global North scene regions higher than Global South regions: all six broad Global North micro-regions appeared in the upper eleven ranks, and nine of the ten lowest-sentiment micro-regions were broad Global South (Supplementary Table~S6). The division was not absolute, however. Five broad Global South regions entered the top ten, including Pacific islands (0.949), Middle Africa (0.915), South-eastern Asia (0.907), Central America (0.902) and Eastern Africa (0.895). Model-level disagreement was concentrated in the Caribbean, Northern Africa, Central Asia and Western Africa, which showed the widest three-model sentiment ranges. Visual-cluster adjustment preserved the regional pattern while also raising the mean sentiment of several Global South micro-regions, including Eastern Africa (0.895 to 0.915), Western Africa (0.798 to 0.848), Central Asia (0.801 to 0.839) and the Caribbean (0.772 to 0.803; Supplementary Fig.~S5b). These baseline differences show that affective unevenness is not only produced by identity-conditioned self--other shifts; it is also present in the default tone assigned to scenes from different parts of the world. Taken together, these results show two related but distinct forms of cultural unevenness in open-ended urban perception: a semantic baseline closer to some regional lenses than others, and an affective baseline whose positivity varies across the scenes being described.

\clearpage
\clearpage

\begin{figure}[H]
  \centering
  \includegraphics[width=\textwidth,height=0.775\textheight,keepaspectratio]{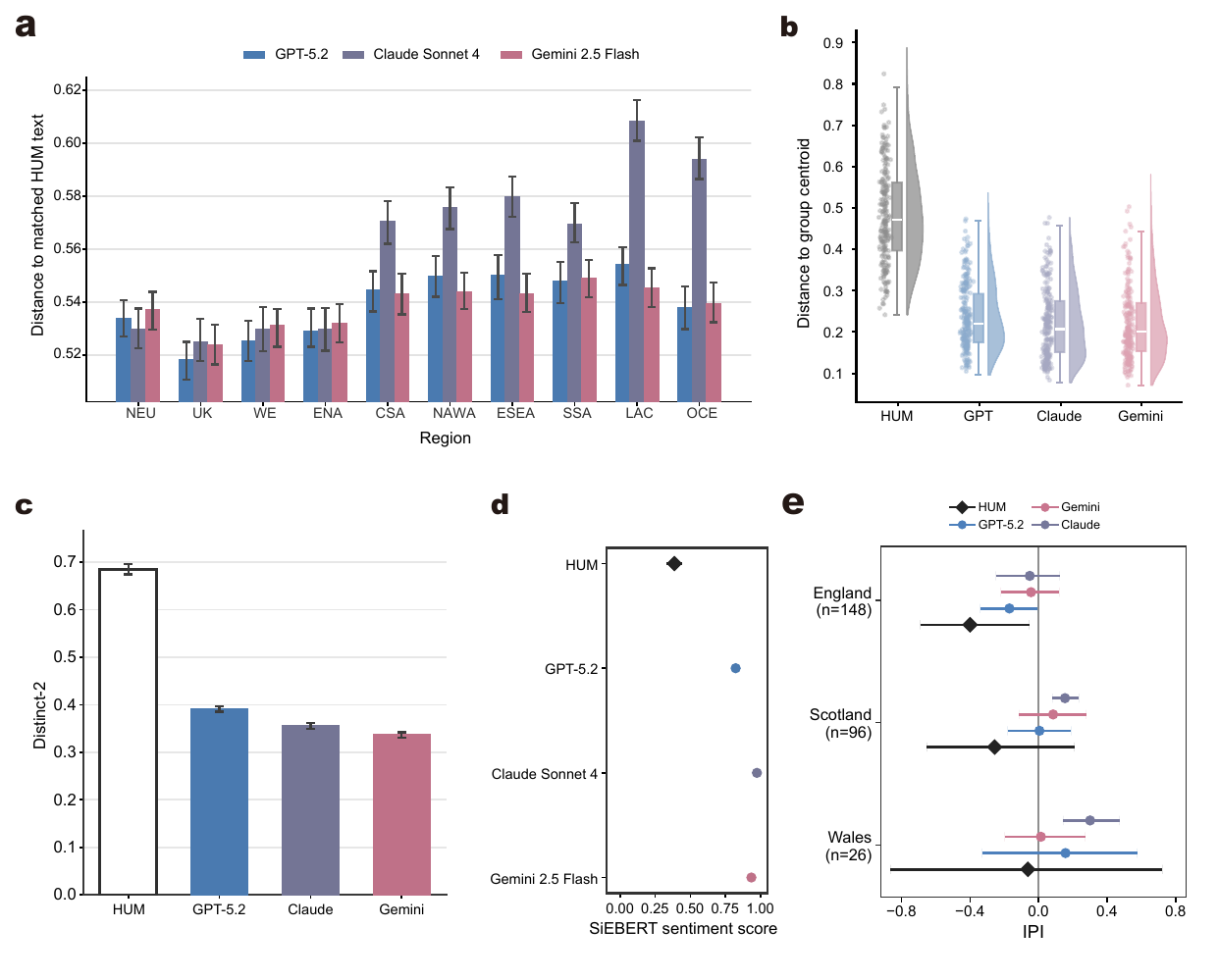}
  \caption{\textbf{UK validation based on Geograph human text-image pairs shows improved contextual alignment but persistent model compression.} a, Semantic distance between model outputs and the matched human text on the 1,000-image Geograph benchmark under neutral, UK, regional-neighbor and global Meso7 prompts. b, Within-group distance to centroid for human texts and model outputs, showing lower dispersion in model-generated language. c, DISTINCT-2 diversity scores for the human texts and each model. d, Sentiment scores derived from the Geograph benchmark texts and the corresponding model outputs. e, Nation-level IPI estimates for England, Scotland and Wales on the 270-image nation-labeled Geograph subset.}
  \label{fig:fig3}
\end{figure}

The systematic comparison between LLM-based urban perception and regional human open-text perception used human place descriptions from Geograph Britain and Ireland, a crowdsourced archive that aims to collect geographically representative photographs and information for each square kilometre of Great Britain and Ireland. On the 1,000-image comparison set, semantic distance to the matched human text was lowest under prompts that were closer to the UK context, especially the UK prompt itself (Fig.~\ref{fig:fig3}a). Across models, the UK prompt reduced mean semantic distance from the neutral condition by 0.005 to 0.016 cosine-distance units, with UK-prompt distances tightly clustered between 0.518 and 0.525. The reduction was consistent across models, but small in magnitude.

That improvement did not restore human diversity. Human descriptions were substantially more dispersed in semantic space than model outputs, with a mean distance to centroid of 0.734, compared with only 0.391--0.418 for neutral model outputs (Fig.~\ref{fig:fig3}b). The same compression appeared lexically: DISTINCT-2 was 0.686 for the human texts but only 0.337--0.392 for the models (Fig.~\ref{fig:fig3}c). The embedding projection in Supplementary Fig.~S8a shows that model and human texts occupy related semantic regions, with limited overlap rather than complete separation. The diversity gap appears more clearly in the resampling and redundancy summaries: human texts retain higher lexical variety, whereas model texts show higher nearest-neighbour similarity (Supplementary Fig.~S8b,c). Models were also much more positive overall. Human Geograph texts had a mean sentiment score of 0.387, whereas the corresponding model outputs ranged from 0.823 to 0.975 (Fig.~\ref{fig:fig3}d). The models therefore describe places more uniformly and evaluate them more positively.

The same tendency reappears in the nation-level IPI comparison based on the 270-image Geograph subset for England, Scotland and Wales. Human texts produced negative IPIs across all three nations (-0.400 for England, -0.256 for Scotland and -0.062 for Wales), whereas the model estimates were systematically more elevated, and in several cases positive (Fig.~\ref{fig:fig3}e). The upward displacement was visible in all three models and was most pronounced for Claude Sonnet 4. Culturally proximate prompting therefore improved semantic alignment with human descriptions, but the model texts remained less diverse, more affectively positive and more strongly shifted in identity-linked evaluation than the human benchmark.

\subsection*{Study 2 | Structured urban perception}

\begin{figure}[H]
  \centering
  \includegraphics[width=\textwidth]{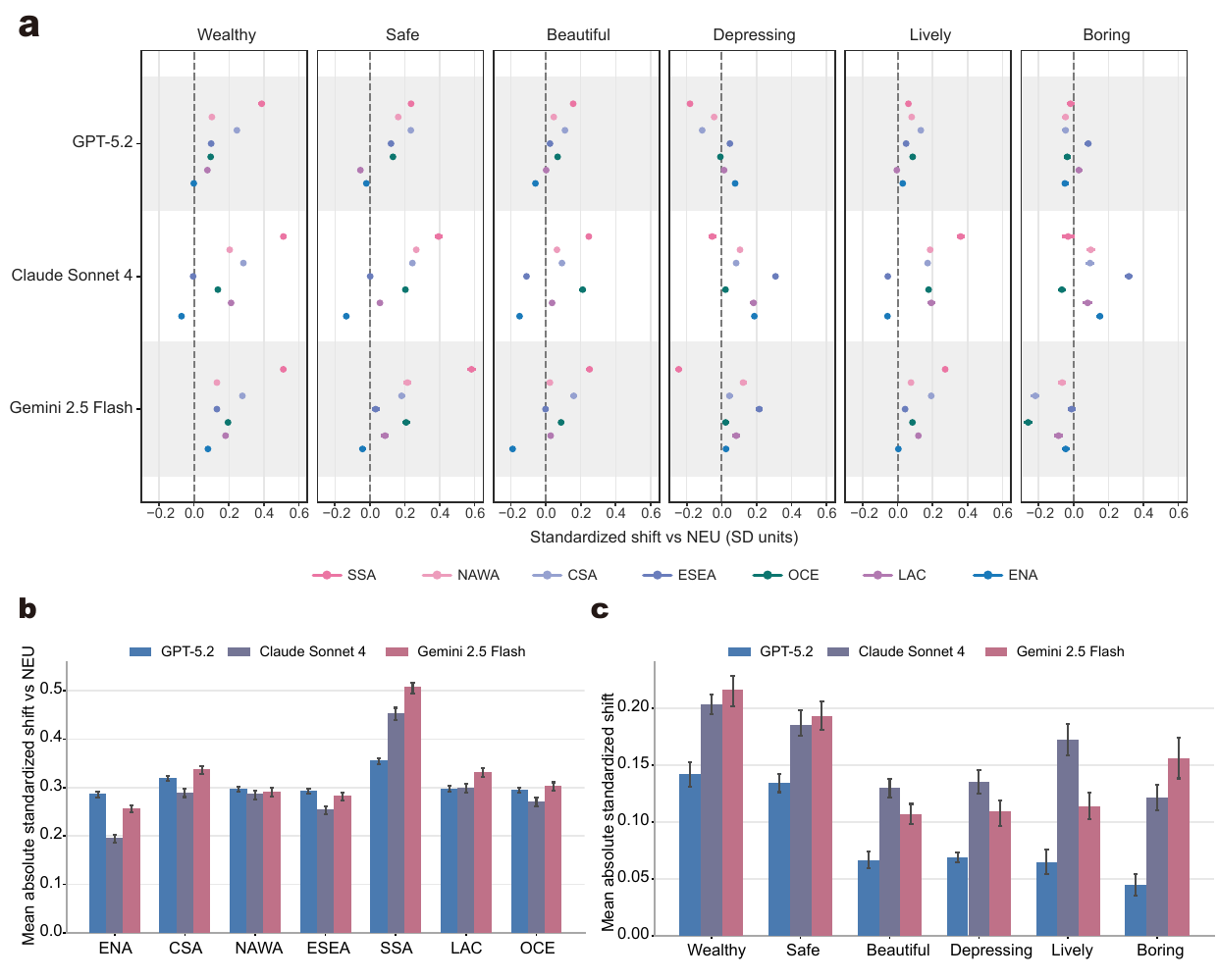}
  \caption{\textbf{Structured six-dimension judgments recover the same culturally patterned shifts seen in open perception.} a, Standardized shifts from the neutral condition for each regional prompt across six perception dimensions and three models. b, Mean absolute standardized shift from neutral by region, aggregated across dimensions for each model. c, Mean absolute standardized shift from neutral by dimension, aggregated across regional prompts for each model.}
  \label{fig:fig4}
\end{figure}

The same cultural structure remained visible when urban perception was measured through a standardized machine-learning framework grounded in human urban-perception judgments\cite{salesses2013collaborative,dubey2016deep}. Applying this perception model to the global street-view sample, we estimated how identity-conditioned framing shifted inferred place evaluations relative to the neutral baseline across six dimensions: wealth, safety, beauty, depression, liveliness and boredom (Fig.~\ref{fig:fig4}a). The resulting shifts were substantial and highly structured rather than random.

The regional distribution of these shifts showed a familiar asymmetry. When the absolute magnitude of identity-conditioned change was aggregated across dimensions, ENA consistently showed the smallest overall displacement from neutral in all three models (Fig.~\ref{fig:fig4}b), ranging from 0.194 to 0.285 s.d. By contrast, SSA produced the largest shifts in every model, ranging from 0.354 to 0.506 s.d. Other non-Western identities also generated larger departures from neutrality, including CSA (0.289--0.335 s.d.) and LAC (0.297--0.330 s.d.). The same ordering was visible when prompt granularity was changed: ENA was the closest Macro5 framing to the structured baseline, while Australia and New Zealand, Northern America and Western Europe formed the lowest-shift group among Micro20 framings (Supplementary Fig.~S10b,c). Thus, the asymmetry seen in the semantic analysis reappears in the structured perception pipeline: the neutral reference frame remains more closely aligned with framings associated with Europe, North America and Australia/New Zealand than with most other prompted identities.

The strength of the effect also depended on which aspect of urban perception was being measured. Across models, perceived wealth was the most identity-sensitive dimension, with mean absolute shifts of 0.141--0.215 s.d., followed by safety at 0.134--0.193 s.d. (Fig.~\ref{fig:fig4}c). Beauty was comparatively more stable at 0.066--0.106 s.d., whereas boredom ranged more widely from 0.044 to 0.155 s.d. This dimension-specific pattern suggests that cultural bias in LLM-based place perception is not uniform: some judgments, especially wealth and safety, are more readily reframed by identity cues than others. The same structured-score sensitivity was also visible when the prompt granularity was changed to Macro5 or Micro20, supporting the view that the structured effect is not an artefact of the main seven-region scheme (Supplementary Fig.~S10). The structured scores thus recover the same broad asymmetry seen in open-text perception, while showing that cultural identity changes some urban-perception dimensions more strongly than others.

\begin{figure}[H]
  \centering
  \includegraphics[width=\textwidth]{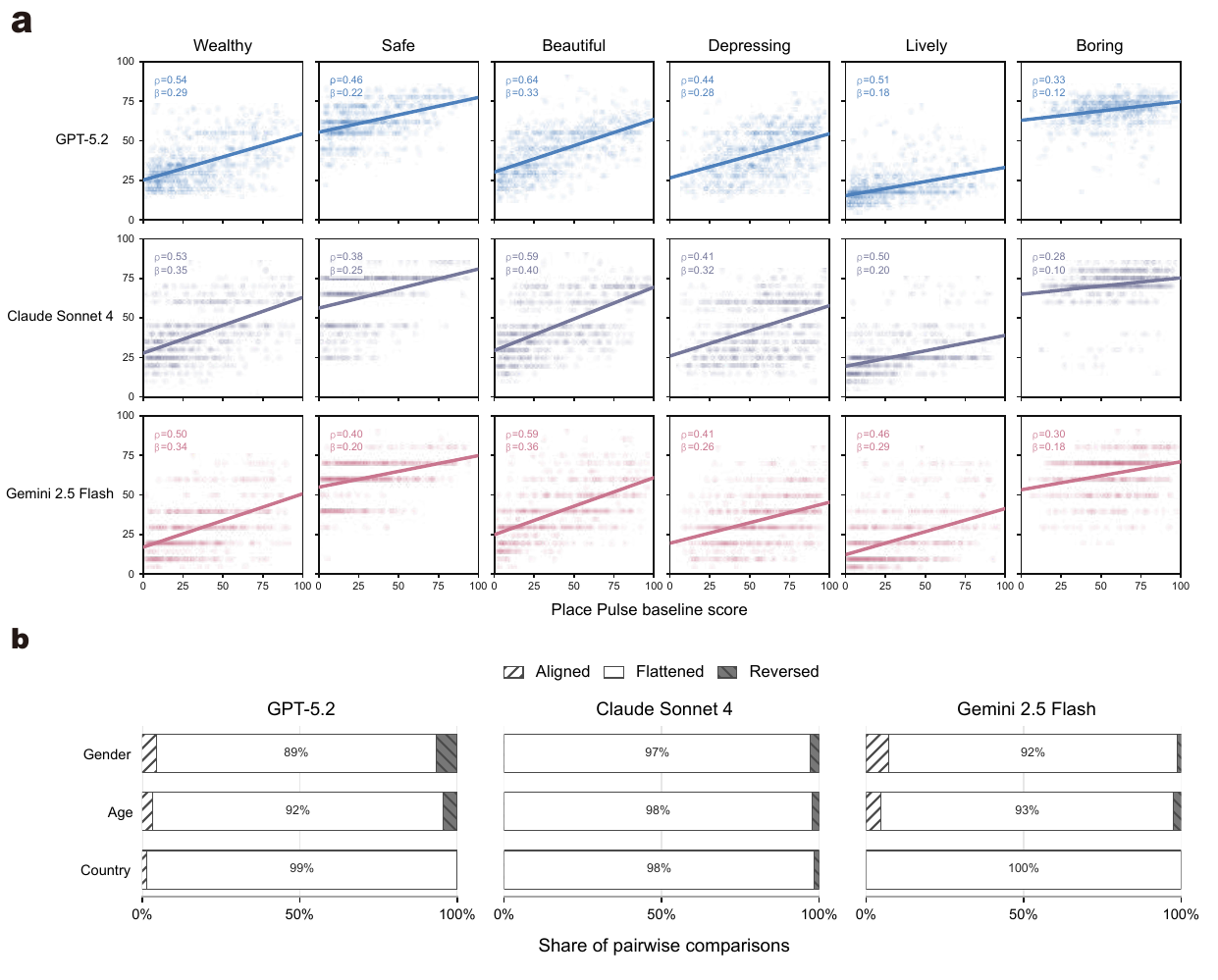}
  \caption{\textbf{External grounding and pairwise subgroup-bias comparison show that structured scores are interpretable but mostly flatten human differences.} a, Dimension-wise relationships between neutral model scores and external baseline scores for wealthy, safe, beautiful, depressing, lively and boring judgments across the three models. Each panel reports Spearman's $\rho$ and the fitted linear slope $\beta$. b, Outcome composition for pairwise gender, age and country comparisons: aligned outcomes reproduce the human subgroup direction, flattened outcomes remove directional subgroup differentiation, and reversed outcomes point in the opposite direction. Mixed or partial cases are retained in the source data but omitted from the stacked bars to focus on the three interpretable response modes.}
  \label{fig:fig5}
\end{figure}

Finally, we compared the LLM-based structured judgments with an existing machine-learning perception model trained on human urban-perception data, and then asked whether the models could reproduce human differences across social categories. Across all six dimensions, LLM and ML scores were positively related, but the strength of agreement varied by perceptual attribute (Fig.~\ref{fig:fig5}a). The closest alignment was observed for beauty, with Spearman correlations of 0.593--0.637, followed by wealth at 0.503--0.541. Agreement was weaker for safety and depression, and weakest for boredom at 0.283--0.325. Thus, the two systems did not produce identical scores, but they were not unrelated either: LLM-based judgments retained a measurable correspondence with the ML perception model, especially on visually salient evaluative dimensions such as beauty and wealth. Across identity-conditioned structured scores, the strongest associations with the external baseline were concentrated in ENA and OCE rather than being evenly distributed across regional framings (Supplementary Fig.~S9).

Beyond aggregate perception scores, a stricter comparison examined whether the models reproduced the direction of subgroup biases observed in human perception data. Pairwise comparisons were constructed for gender, age and country axes, and each model outcome was classified as aligned with the human subgroup direction, flattened, or reversed (Fig.~\ref{fig:fig5}b). The dominant pattern was not reversal but compression. Across all pair-model cases, 81.7\% were flattened, whereas exact alignment and reversal each accounted for only 2.2\%. In the three displayed outcome classes, flattened outcomes dominated every subgroup axis and model, ranging from 88.9\% to 100\% across the Fig.~\ref{fig:fig5}b summaries. Country comparisons were especially unlikely to show directional alignment, with aligned outcomes at 0--1.5\% among the displayed classes. The same conclusion holds when the outcome composition is broken down by perception dimension: flattened outcomes dominate the displayed classes across safety, beauty, wealth, liveliness, boredom and depression (Supplementary Fig.~S11).

These structured comparisons therefore indicate two limits of LLM-based urban perception. First, model scores retain measurable correspondence with an existing human-perception model, but only unevenly across dimensions. Second, when human subgroup differences are used as the reference, the models more often erase directional subgroup structure than reproduce it.

\section*{Discussion}

AI bias becomes especially consequential when models are used to interpret cities. Urban perception has long been a way of translating physical environments into social meaning, from early work on the image and evaluation of the city to contemporary large-scale perception mapping\cite{nasar1990evaluative,salesses2013collaborative}. Multimodal models now extend this measurement tradition by generating descriptions and scores directly from urban images. This creates a methodological opportunity for urban analytics and planning, but it also changes the object being measured. When AI systems are used to describe, rank or compare urban environments, their outputs may carry cultural assumptions about what a city should look and feel like.

Our findings show that this risk is systematic. Across open-ended descriptions, affective language and structured place judgments, the general condition behaved as a culturally situated reference point. Responses generated under Europe and Northern America framing were systematically closer to this general condition than responses generated under other regional framings, and the same broad asymmetry reappeared in structured scores. Europe and Northern America should therefore be understood as a diagnostic position in the model's perceptual space, not as a ground truth for urban perception. The broader implication is that bias in AI-based urban perception is not only a deviation from a neutral output. The reference condition used to define neutrality can itself encode a cultural location. In this study, the models translated visible urban form through uneven cultural priors, producing systematic shifts across description, affect and evaluation.

The comparison with human descriptions further shows why contextual adjustment is not enough. Regionally proximate framing, especially the UK framing in the Geograph comparison, moved model descriptions modestly closer to local human descriptions. Yet this movement did not recover the heterogeneity of human place perception. Human texts were more dispersed in semantic space, more lexically varied and less uniformly positive than model outputs for the same scenes. The models therefore moved toward a regional human reference without reproducing the diversity of human responses. This distinction matters because local perception is not simply a regional label attached to an image. It is produced through situated experience, selective attention, memory, mobility, neighbourhood familiarity, social position and disagreement. A model can generate a fluent regional voice, but this should not be treated as lived perception. The risk is proxy perception: AI outputs may appear locally sensitive while still compressing the plurality of the publics they implicitly represent. This echoes broader evidence that LLMs used as substitutes for human participants can misportray and flatten identity groups, as well as concerns that cultural variation in model outputs can be simplified or erased\cite{wang2025flatten,qadri2025erasure}.

The structured task raises a parallel issue about the validity of AI-generated urban indicators. Neutral model scores were positively associated with Place Pulse-style perception baselines, especially for visually salient dimensions such as beauty and wealth. The scores were therefore not arbitrary: they recovered recognizable axes of urban perception and could plausibly be used to rank, map or compare urban scenes. However, predictive coherence is not the same as representational validity. Perceived safety, beauty, liveliness, wealth, boredom and depression are social judgments, not objective physical measurements. Their validity depends not only on whether a model reproduces an average score, but also on whether it preserves meaningful variation across the people who perceive the city. Recent global evidence shows that urban visual perception varies across demographics and personalities rather than forming a single universal standard\cite{quintana2025global}. In our subgroup comparison, the dominant model outcome was flattening, where subgroup differences observed in human perception were removed rather than reproduced. A model-generated map of perceived safety or beauty may therefore look stable, scalable and policy-relevant while replacing plural human perception with a compressed model view. Such a map can be attractive precisely because it removes disagreement. But in urban planning and design, disagreement is not noise; it is part of what urban perception is.

These findings have direct implications for urban analytics, planning support and AI-assisted design. Multimodal models are likely to be used for scene summarization, neighbourhood diagnosis, design precedent search, participatory synthesis, environmental auditing and policy support\cite{fu2025llmplanning,batty2025generative}. In these settings, cultural non-neutrality is not a narrow benchmarking issue. It can influence which streets are described as orderly or neglected, which public spaces are framed as lively or unsafe, which forms of density are treated as attractive or problematic, and which local concerns are translated into general urban knowledge. Auditing urban AI should therefore go beyond aggregate accuracy. It should evaluate the cultural location of the general reference condition, the affective tone of generated descriptions, the sensitivity of structured scores to cultural framing, and the degree to which models preserve or suppress subgroup differences. Calibration against local data may help, but calibration should not be treated as a purely technical correction. Human benchmarks themselves carry geographic, demographic and historical structure, and should be used as situated empirical references rather than universal ground truth.

Several limitations remain. We study three contemporary multimodal models, and the model landscape will continue to change. The open-text analysis is conducted in English, using English instructions and English-language semantic and sentiment tools; multilingual evaluation and locally authored criteria may reveal different forms of alignment or misalignment. We focus on street-view imagery rather than the full multisensory, temporal and social experience of urban life. Regional cultural framings are necessarily stylized and cannot represent the diversity within regions, countries, cities or neighbourhoods. This is also why role-based model responses should not be equated with genuine standpoint-taking\cite{shanahan2023roleplay}. The human comparison data are geographically uneven: Geograph provides a useful text-image benchmark for Britain and Ireland, while the external qscore summaries provide subgroup references for structured judgments, but neither covers the full geography of the global street-view sample. Future work should compare model outputs against multiple human publics rather than a single aggregate benchmark, and examine whether calibration can preserve disagreement rather than merely improve average fit. The central conclusion is that AI-based urban perception is not a neutral substitute for human perception. It is shaped by culturally organized priors, and those priors should be treated as part of the urban evidence that AI systems produce.

\section*{Methods}

\subsection*{Global image corpus and analysis sample selection}

We assembled a merged street-view corpus from multiple providers while retaining explicit source provenance, image dates and copyright metadata. Candidate urban locations were identified through spatial anchor layers and related discovery procedures, but these locations were used only for discovery rather than as final analytical units. For Google Street View imagery, panorama availability was first checked through the metadata service of the Street View Static API and images were then retrieved through the corresponding image endpoint. Equivalent provenance information was retained for all sources, including Baidu Panorama where required, so that image origin remained auditable throughout the workflow. The 3,000-scene analytical sample used in the main figures was drawn directly from this larger merged corpus rather than from the anchor layer itself.

From the merged corpus we selected a scene-balanced global street-view image set of 3,000 scenes. Before sampling, images were embedded to derive coarse visual clusters, and we ran a soft indoor-screening audit to reduce obvious non-street scenes. Sampling quotas were then set across visual cluster, place type, country and provider to prevent the main analysis from being dominated by a small number of recurrent scene types or a narrow geographic subset. It also followed an 8:2 split between locations classified by the GHS framework as urban centre or town settings, preserving both geographic and visual diversity\cite{durmanual2021,ghssmod2023}. Regional coverage was tracked with a UN-based geographic hierarchy implemented on Natural Earth Admin-0 country polygons\cite{naturalearth2026admin0,unsd2026m49}. In the country table used for sampling, these three levels correspond to Macro5 (UN Region), Micro20 (UN Subregion) and Meso7 (UN SDG). The spatial distribution and region-level counts of the final analysis set are shown in Supplementary Fig.~S1, and the mapping used in the manuscript is summarized in Supplementary Table~S4. A separate 500-image subset was reserved for robustness checks and was not used in the main 3,000-scene figures.

\subsection*{Open-perception prompting and three-model generation}

For Study 1, each image in the global street-view image set was evaluated under eight prompt conditions: one neutral condition and seven Meso7 cultural contexts (ENA, CSA, NAWA, ESEA, SSA, LAC and OCE; full names in Supplementary Table~S5). These Meso7 identities follow the UN SDG regional grouping used in SDG reporting (Supplementary Table~S4). Prompts asked the model to produce a short English paragraph grounded in visible scene evidence. The three-model comparison used GPT-5.2, Claude Sonnet 4 and Gemini 2.5 Flash. Because all 3,000 images were evaluated under all eight conditions for all three models, the open-perception analysis comprises 72,000 model-generated texts.

\subsection*{Semantic embedding, neutral-distance, and ingroup-preference analysis}

Open-text outputs were embedded with the fixed sentence encoder \texttt{all-mpnet-base-v2}. Let $t_{imc}$ denote the text generated for image $i$ by model $m$ under prompt condition $c$, and let $e_{imc}=f_{\text{embed}}(t_{imc})$ be its embedding. For every non-neutral condition, we measured semantic displacement from the matched neutral response for the same image:
\[
d_{imc}=1-\cos\!\left(e_{im,\mathrm{NEU}},e_{imc}\right)
\]
where larger values indicate greater prompt-induced semantic movement. Region-level neutral distance was then summarized as
\[
D_{mc}=\frac{1}{N}\sum_{i=1}^{N} d_{imc}.
\]
To visualize the geometry of prompt conditions, we computed condition centroids in embedding space and projected them with local low-dimensional ordination. We also quantified within-condition dispersion relative to the centroid,
\[
\delta_{mc}=\frac{1}{N}\sum_{i=1}^{N}\left(1-\cos\!\left(e_{imc},\bar e_{mc}\right)\right),
\]
where $\bar e_{mc}$ is the centroid embedding for model $m$ and condition $c$.

Sentiment was estimated from the same texts with SiEBERT (\texttt{sentiment-roberta-large-english}). For each text, the classifier returns positive and negative posterior probabilities, denoted $p^{(+)}_{imc}$ and $p^{(-)}_{imc}$. We define the sentiment score as
\[
s_{imc}=p^{(+)}_{imc}-p^{(-)}_{imc},
\]
so that higher values indicate more positive place evaluation. Region-level ingroup preference indices (IPIs) were then computed by comparing the sentiment assigned by a region-matched prompt to scenes from that region against the mean sentiment assigned by non-matching prompts to the same scenes:
\[
\mathrm{IPI}_{mr}=\frac{\mu^{\mathrm{self}}_{mr}-\mu^{\mathrm{other}}_{mr}}{\sigma_{mr}},
\]
where $\mu^{\mathrm{self}}_{mr}$ is the mean sentiment for images from region $r$ under the matching regional prompt, $\mu^{\mathrm{other}}_{mr}$ is the corresponding mean under non-matching prompts, and $\sigma_{mr}$ is the within-region standard deviation of sentiment scores.

For Fig.~\ref{fig:fig2}c, uncertainty in IPI estimates was obtained by bootstrapping images within each region and recomputing the self--other contrast. For Fig.~\ref{fig:fig2}d, we summarized neutral-prompt sentiment by scene micro-region and model, again using image-level bootstrap intervals. Micro-regions were ordered by their mean neutral sentiment across the three models so that the panel shows geographic variation in the affective baseline rather than prompt-specific ingroup preference.

\subsection*{Geograph-based UK benchmark construction}

All UK analyses were based on Geograph human text-image pairs. We constructed a visually grounded benchmark subset by retaining entries whose accompanying human text described the visible scene with sufficient specificity for comparison to model outputs, while also capping repeated contributions from the same author to reduce dominance by a small number of prolific contributors. This yielded a 1,000-image Geograph benchmark subset for open-text comparison. From the same source, we then derived a separate 270-image subset with nation labels for England, Scotland and Wales, allowing nation-level comparison without introducing a second human data source.

\subsection*{Open-text comparison with Geograph human texts}

On the 1,000-image Geograph benchmark, we compared model outputs to the matched human description embedded in the same semantic space. If $h_i$ denotes the human text paired with image $i$, with embedding $e^{\mathrm{hum}}_i$, then the model-to-human semantic distance is
\[
h_{imc}=1-\cos\!\left(e^{\mathrm{hum}}_i,e_{imc}\right).
\]
For each model and condition, we summarize the benchmark by the mean human-distance $\bar h_{mc}$ and by the improvement relative to neutral,
\[
\Delta^{\mathrm{hum}}_{mc}=\bar h_{mc}-\bar h_{m,\mathrm{NEU}},
\]
where negative values indicate better alignment with the human text than the neutral prompt. The UK comparison combines a core set of culturally proximate prompts (neutral, UK, Western Europe and ENA) with additional Meso7 prompts used to place the UK benchmark back into the broader global prompt frame. We report four complementary summaries: semantic distance from the matched human text, within-group distance to centroid, DISTINCT-2 lexical diversity, and sentiment score. These measures test both contextual proximity to human language and the extent to which model outputs recover human heterogeneity.

\subsection*{Nation-level IPI comparison on Geograph subsets}

For the 270-image nation-labeled Geograph subset, we prompted the models under England, Scotland and Wales conditions and compared the resulting nation-level IPIs to the human baseline derived from the Geograph texts. The nation-level calculation follows the same formula as above, with nations in place of meso-regions. The purpose of this comparison was not to re-estimate the full UK benchmark, but to test whether the models preserved within-UK differentiation rather than collapsing the country into a single perceptual identity.

\subsection*{Structured six-dimension prompting}

For Study 2, the same global street-view image set was evaluated again under the same eight prompt conditions, but the response format was constrained to six explicit dimensions: \texttt{safe}, \texttt{lively}, \texttt{wealthy}, \texttt{beautiful}, \texttt{boring}, and \texttt{depressing}. Outputs were returned as structured scores on a common 0--100 scale. Let $y_{imck}$ denote the score for image $i$, model $m$, condition $c$, and dimension $k$. We first compute the within-image shift from the neutral prompt,
\[
\Delta y_{imck}=y_{imck}-y_{im,\mathrm{NEU},k},
\]
and then summarize each region-by-dimension effect as a standardized mean shift,
\[
z_{mck}=\frac{\frac{1}{N}\sum_{i=1}^{N}\Delta y_{imck}}{\mathrm{sd}_i\!\left(y_{im,\mathrm{NEU},k}\right)}.
\]
These standardized shifts are the values plotted in Fig.~\ref{fig:fig4}a. We also aggregate $\lvert z_{mck}\rvert$ by region and by perceptual dimension to quantify which prompts and which judgments are most sensitive to cultural reframing.

\subsection*{External perception grounding analysis}

To test whether the neutral structured scores remained connected to established urban perception benchmarks, we compared them against baseline scores produced by an open-source human-perception model trained on large-scale pairwise urban-image judgments and released with the Global Streetscapes project\cite{salesses2013collaborative,dubey2016deep,globalstreetscapes2024}. We used the six dimension-specific checkpoints for safety, lively, wealthy, beautiful, boring and depressing judgments. The released model uses a ViT-B/16 backbone with a task-specific multilayer classification head, yielding one baseline score $p_{ik}$ for image $i$ and perceptual dimension $k$. For each dimension $k$ and model $m$, we estimated the association between the neutral LLM score and the corresponding external baseline score $p_{ik}$ through both a Spearman correlation
\[
\rho_{mk}=\mathrm{corr}_{S}\!\left(y_{im,\mathrm{NEU},k},p_{ik}\right)
\]
and a fitted linear slope from
\[
y_{im,\mathrm{NEU},k}=\alpha_{mk}+\beta_{mk} p_{ik}+\varepsilon_{imk}.
\]
We report both $\rho_{mk}$ and $\beta_{mk}$ because the first captures rank-order agreement and the second captures directional scaling. As a supplementary extension, we repeated the same correlation summary after replacing the neutral score $y_{im,\mathrm{NEU},k}$ with each identity-conditioned score $y_{im,c,k}$, where $c$ indexes the seven regional framings and the neutral condition.

\subsection*{External pairwise subgroup-bias comparison}

The final analysis module used SPECS qscore summaries from the Nature Cities study by Quintana et al.\cite{quintana2025global} to ask whether LLM structured judgments reproduced the direction of subgroup differences observed in human street-scene perception data. We constructed pairwise street-scene comparisons for three subgroup axes, gender, age and country, with 90 candidate pairs per axis and 15 candidate pairs per perceptual dimension. Each pair carried subgroup-specific human orderings derived from the corresponding qscore summaries. For a pair $p$, subgroup axis $g$, subgroup $s$, model $m$ and dimension $k$, let $h_{pgsk}$ denote the human-implied expected choice among \{\textit{left}, \textit{right}, \textit{same}\}, and let $\ell_{pgmsk}$ denote the corresponding model choice after comparing its structured scores for the two images.

We classified each pair-level model response into four mutually exclusive outcomes. A pair was \textit{aligned} if all subgroup-specific model choices matched the human-implied choices:
\[
\ell_{pgmsk}=h_{pgsk}\quad\text{for all }s.
\]
It was \textit{reversed} if all subgroup-specific choices pointed in the opposite left-right direction, treating \textit{same} as its own opposite. It was \textit{flattened} if the model returned the same choice for all subgroups in that axis, thereby removing the directional subgroup differentiation present in the human summaries. Remaining cases were labelled \textit{mixed/partial}. Figure~\ref{fig:fig5}b focuses on the three interpretable outcome modes--aligned, flattened and reversed--by subgroup axis; the corresponding breakdown by perception dimension is reported in Supplementary Fig.~S11. Mixed/partial cases are retained in the released source data.

\subsection*{Bootstrap and uncertainty estimation}

For mean distances, score shifts and model-benchmark associations, point estimates were paired with nonparametric 95\% bootstrap confidence intervals, taken as the 2.5th and 97.5th percentiles of the bootstrap distribution. For image-based analyses, bootstrap resampling was performed at the image level while preserving the matched condition structure attached to each image. The same bootstrap framework was used for semantic distances, centroid dispersion, human-alignment deltas, IPIs, structured-score shifts and external-benchmark summaries. The pairwise subgroup-bias analysis in Fig.~\ref{fig:fig5}b and Supplementary Fig.~S11 is reported as observed outcome composition across pair-level classifications rather than as a bootstrap interval estimate.

\section*{Data availability}

Derived source data for the main figures and curated supplementary analyses are available at \url{https://github.com/jameslemon2002/aibias}. Street-view imagery and other provider-restricted visual assets are not redistributed in the public repository.

\section*{Code availability}

Public scripts used to regenerate the released supplementary figures, robustness figures and prompt templates are available at \url{https://github.com/jameslemon2002/aibias}. The repository excludes manuscript source files, API credentials, private request headers and local intermediate caches tied to licensed or non-redistributable inputs.

\bibliographystyle{naturemag}
\bibliography{references}

\clearpage
\clearpage
\section*{Supplementary Information}

\setcounter{figure}{0}
\renewcommand{\thefigure}{S\arabic{figure}}
\renewcommand{\figurename}{Supplementary Figure}
\setcounter{table}{0}
\renewcommand{\thetable}{S\arabic{table}}
\renewcommand{\tablename}{Supplementary Table}

This Supplementary Information file is organized in the order requested for submission: Supplementary Figures, Supplementary Tables and Supplementary Notes. Supplementary Figures are arranged by the order in which the corresponding evidence enters the manuscript: study design and sampling, Study 1 open-text perception and Study 2 structured perception.

\section*{Supplementary Figures}

\subsection*{Study design and sampling}

\begin{figure}[H]
  \centering
  \includegraphics[width=\textwidth,height=0.70\textheight,keepaspectratio]{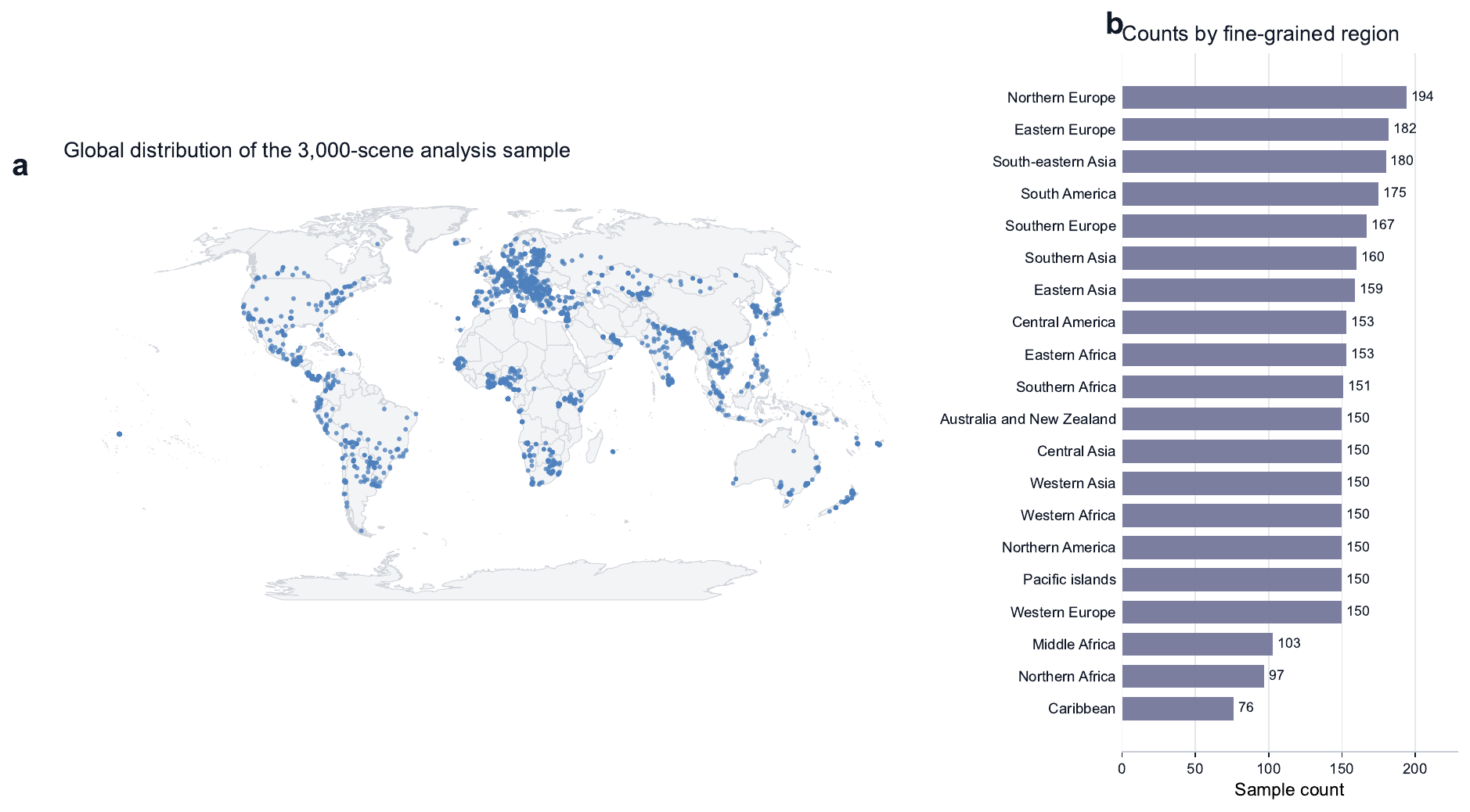}
  \caption{\textbf{Spatial distribution of the 3,000-scene analysis sample.} a, Global distribution of the final analysis scenes selected from the merged street-view corpus. Points show the retained scene locations after balancing across visual cluster, place type, country and provider. b, Sample counts by fine-grained region in the final analysis set. Counts remain close to the nominal target while varying where feasible coverage constraints limited the number of retained scenes.}
\end{figure}

\clearpage
\subsection*{Study 1: open-text perception}

\begin{figure}[H]
  \centering
  \includegraphics[width=\textwidth,height=0.58\textheight,keepaspectratio]{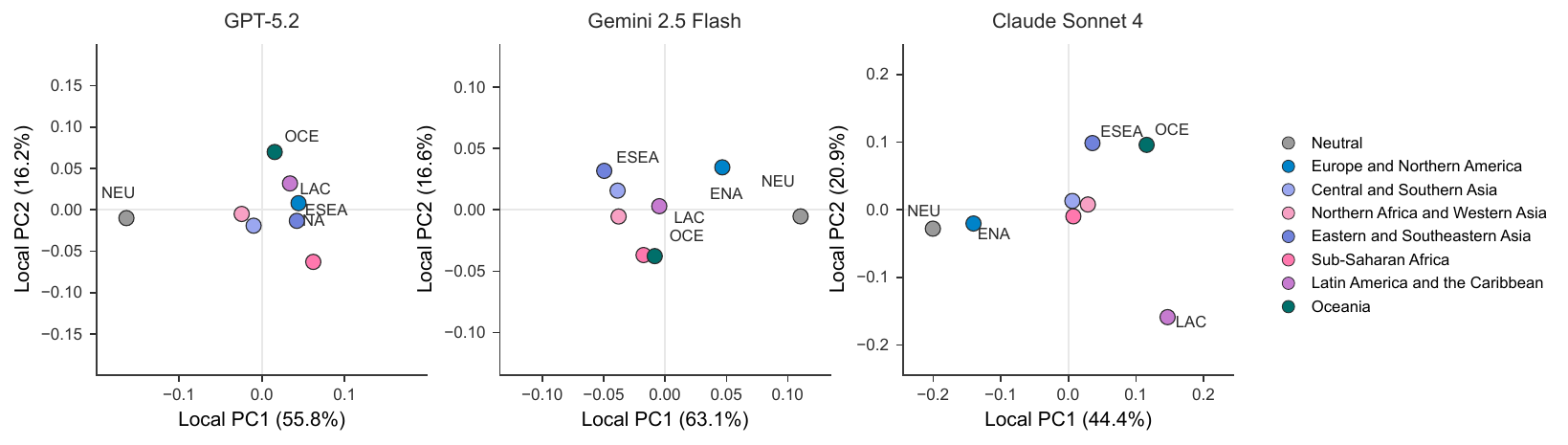}
  \caption{\textbf{Local PCA centroid structure of open-text descriptions.} Points show condition centroids for the neutral prompt and seven regional cultural framings, plotted separately for GPT-5.2, Gemini 2.5 Flash and Claude Sonnet 4. Each panel uses a local two-dimensional PCA fitted within the corresponding model's description embeddings. The centroids occupy distinct positions rather than forming an undifferentiated cloud relative to the neutral condition, indicating that regional framing changes the direction as well as the magnitude of semantic displacement. Axis labels report the variance explained by each local component.}
\end{figure}

\clearpage
\begin{figure}[H]
  \centering
  \includegraphics[width=\textwidth,height=0.68\textheight,keepaspectratio]{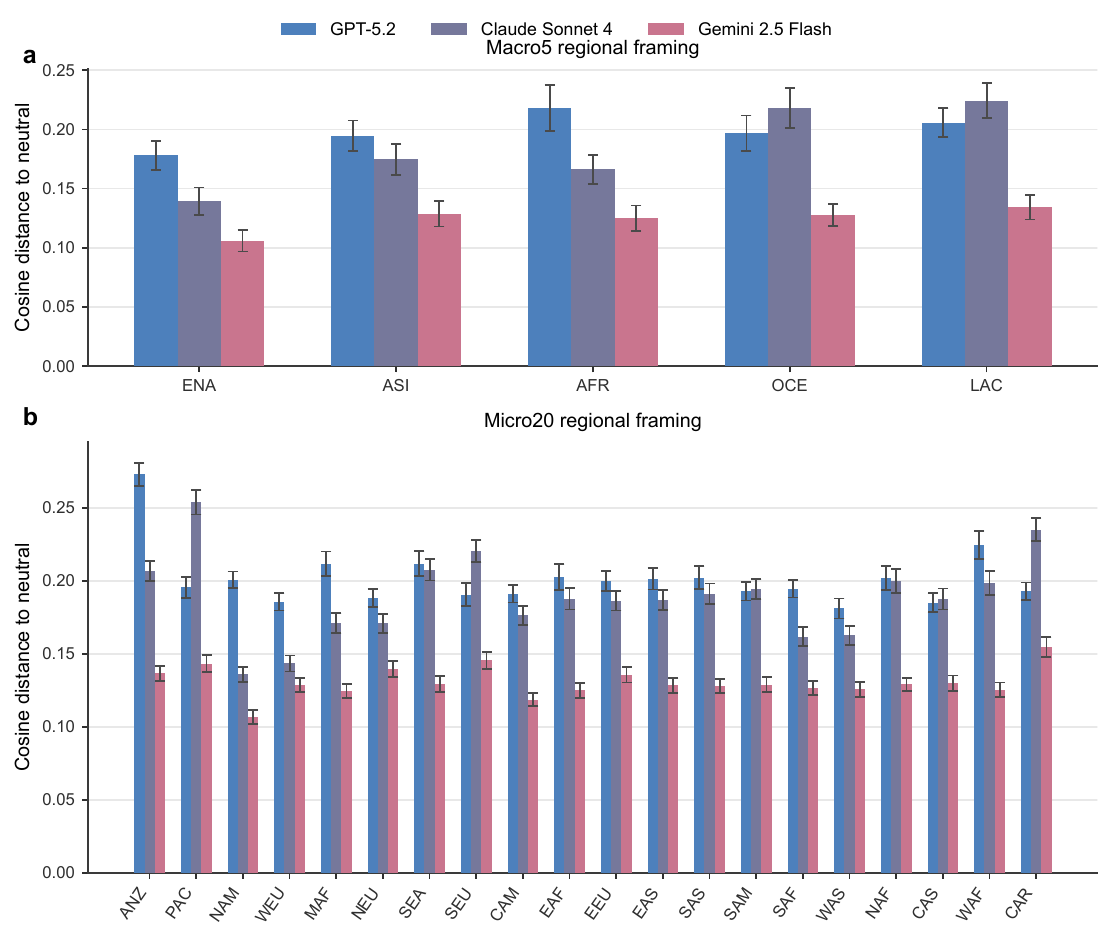}
  \caption{\textbf{Semantic neutral-distance robustness under coarser and finer regional framings.} a, Macro5 regional framings evaluated on an independent 100-image diagnostic subset with 20 images per Macro5 region. b, Micro20 regional framings evaluated on an expanded independent 400-image diagnostic subset with 20 images per Micro20 region. Bars show the mean cosine distance from each identity-conditioned description to the matched general-prompt description, with error bars showing approximate 95\% image-level intervals. The Micro20 panel uses the same abbreviations and ordering as Fig. 2d. ENA remains the closest Macro5 framing, while NAM and WEU are the closest Micro20 framings.}
\end{figure}

\clearpage
\begin{figure}[H]
  \centering
  \includegraphics[width=\textwidth,height=0.73\textheight,keepaspectratio]{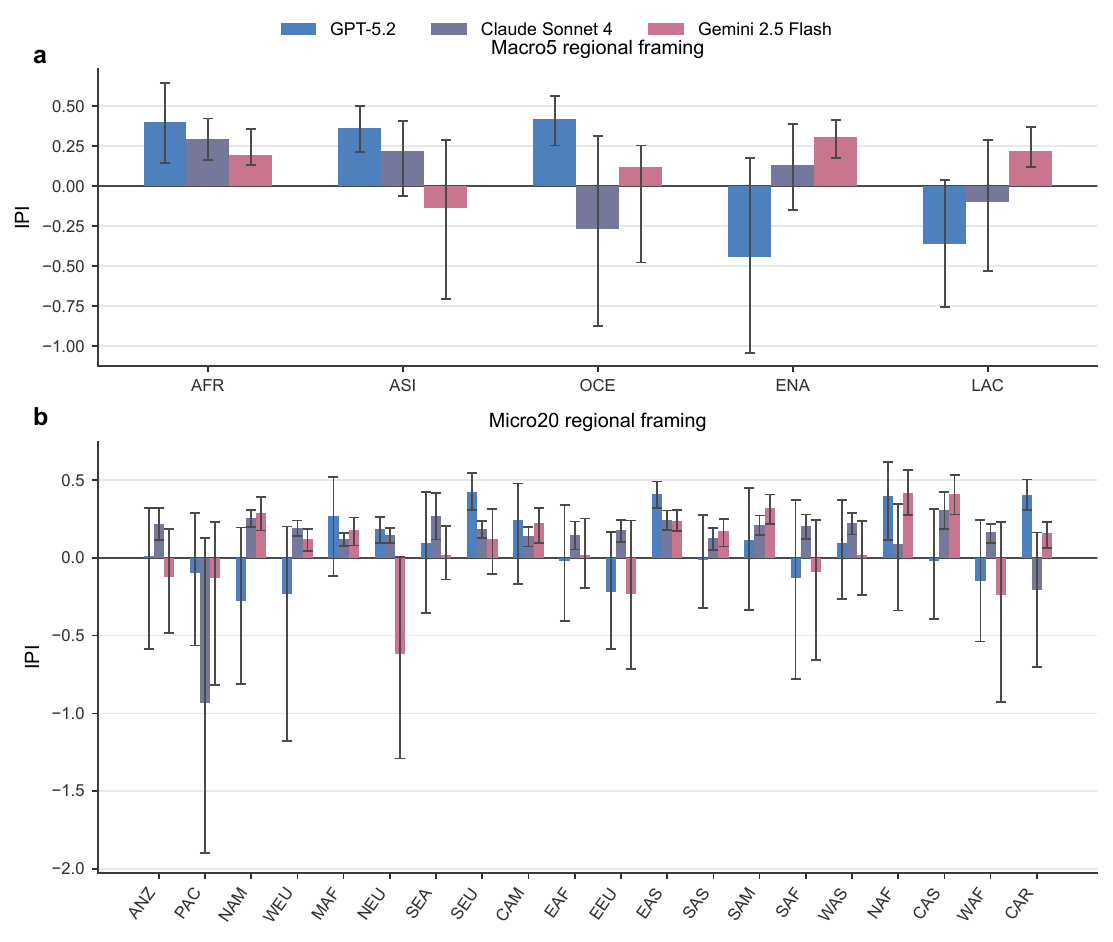}
  \caption{\textbf{Affective ingroup-preference robustness under coarser and finer regional framings.} a, Macro5 sentiment-based ingroup preference index (IPI) evaluated on the independent 100-image diagnostic subset with 20 images per Macro5 region. b, Micro20 IPI evaluated on the expanded independent 400-image diagnostic subset with 20 images per Micro20 region. Bars show model-specific IPI estimates and error bars show 95\% image-level bootstrap intervals. Positive values indicate that an identity-conditioned framing describes scenes from its own region more positively than other identity framings describe the same scenes. The Micro20 panel uses the same abbreviations and ordering as Fig. 2d.}
\end{figure}

\clearpage
\begin{figure}[H]
  \centering
  \includegraphics[width=\textwidth,height=0.78\textheight,keepaspectratio]{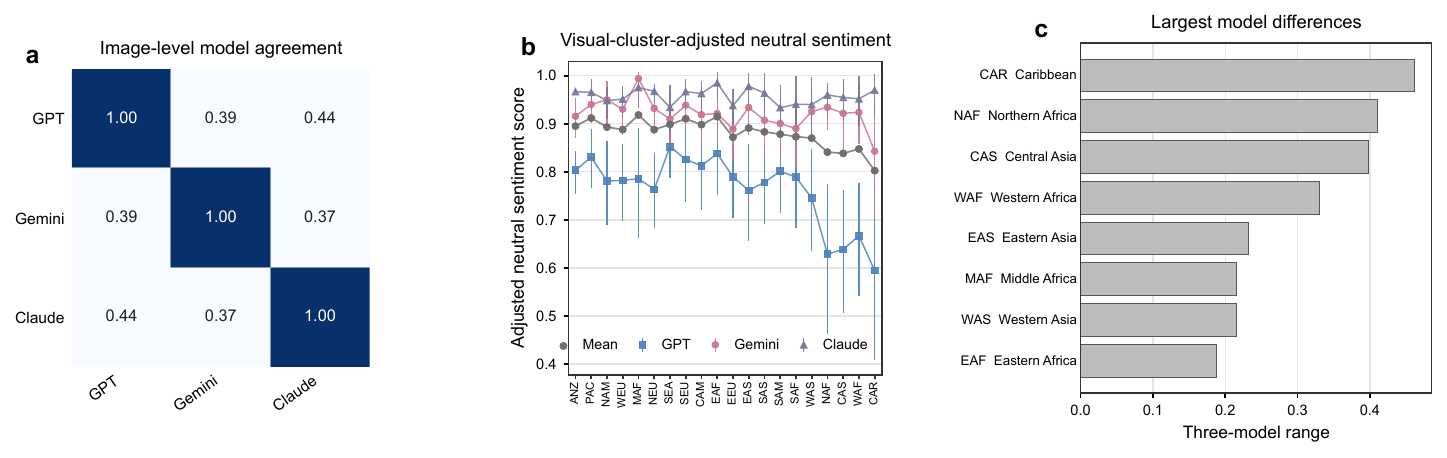}
  \caption{\textbf{Neutral-prompt affective baseline across 20 scene micro-regions.} a, Image-level Spearman correlations in neutral sentiment scores for the same 3,000 street-view scenes across GPT-5.2, Gemini 2.5 Flash and Claude Sonnet 4. b, Region-level neutral sentiment after visual-cluster adjustment. Points and error bars show image-level bootstrap means and 95\% intervals, ordered as in Fig. 2d. c, Micro-regions with the largest three-model range in mean neutral sentiment. These panels characterize the general-prompt affective baseline rather than identity-conditioned IPI.}
\end{figure}

\clearpage
\begin{figure}[H]
  \centering
  \includegraphics[width=\textwidth,height=0.67\textheight,keepaspectratio]{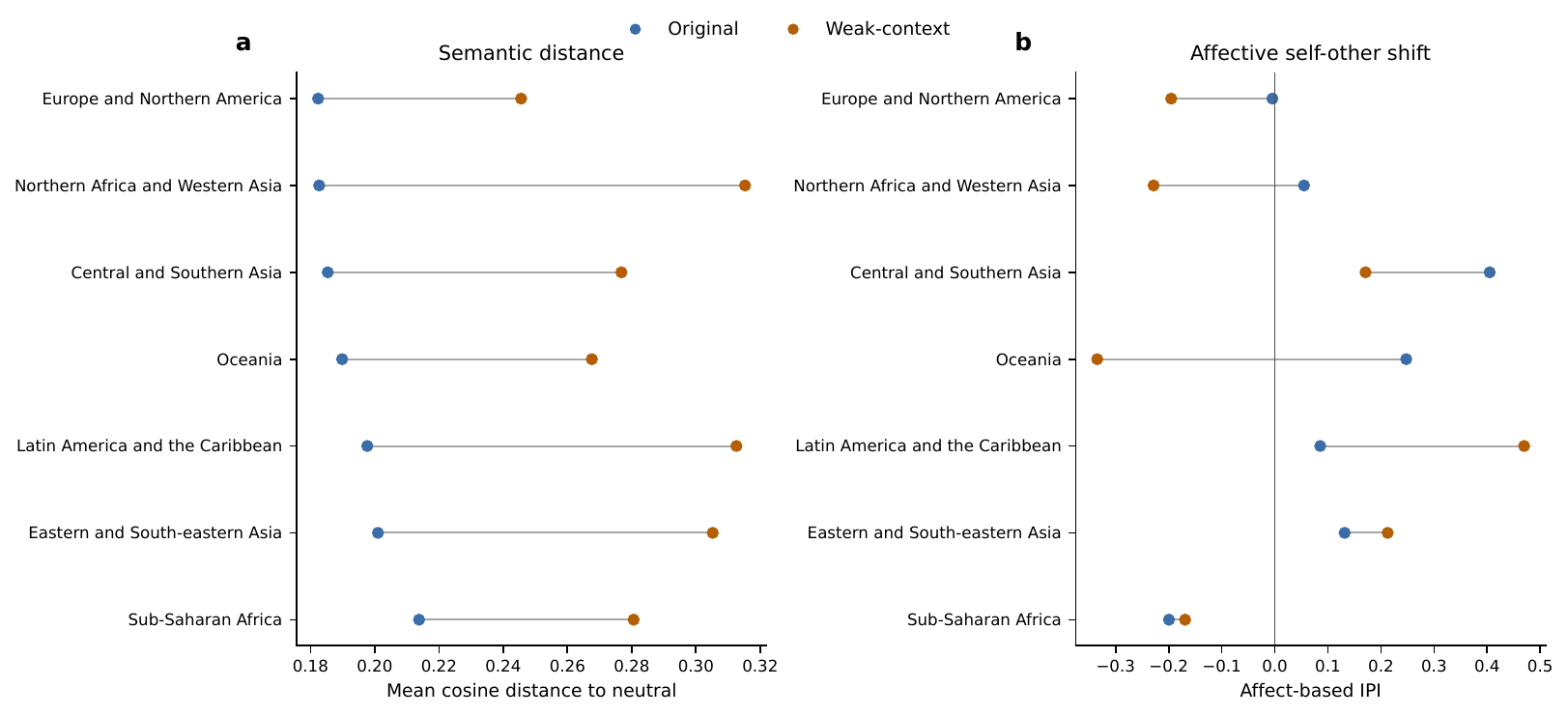}
  \caption{\textbf{Sensitivity of Meso7 effects to a weaker, non-role-playing prompt.} The prompt-form check used an independent 140-image Meso7-balanced diagnostic subset with 20 scenes per Meso7 region. a, Paired semantic distance to the neutral baseline under the original Meso7 prompt and a weaker context-based prompt. b, Paired affect-based IPI under the same two prompt forms. Lines connect estimates for the same regional framing.}
\end{figure}

\clearpage
\begin{figure}[H]
  \centering
  \includegraphics[width=\textwidth,height=0.69\textheight,keepaspectratio]{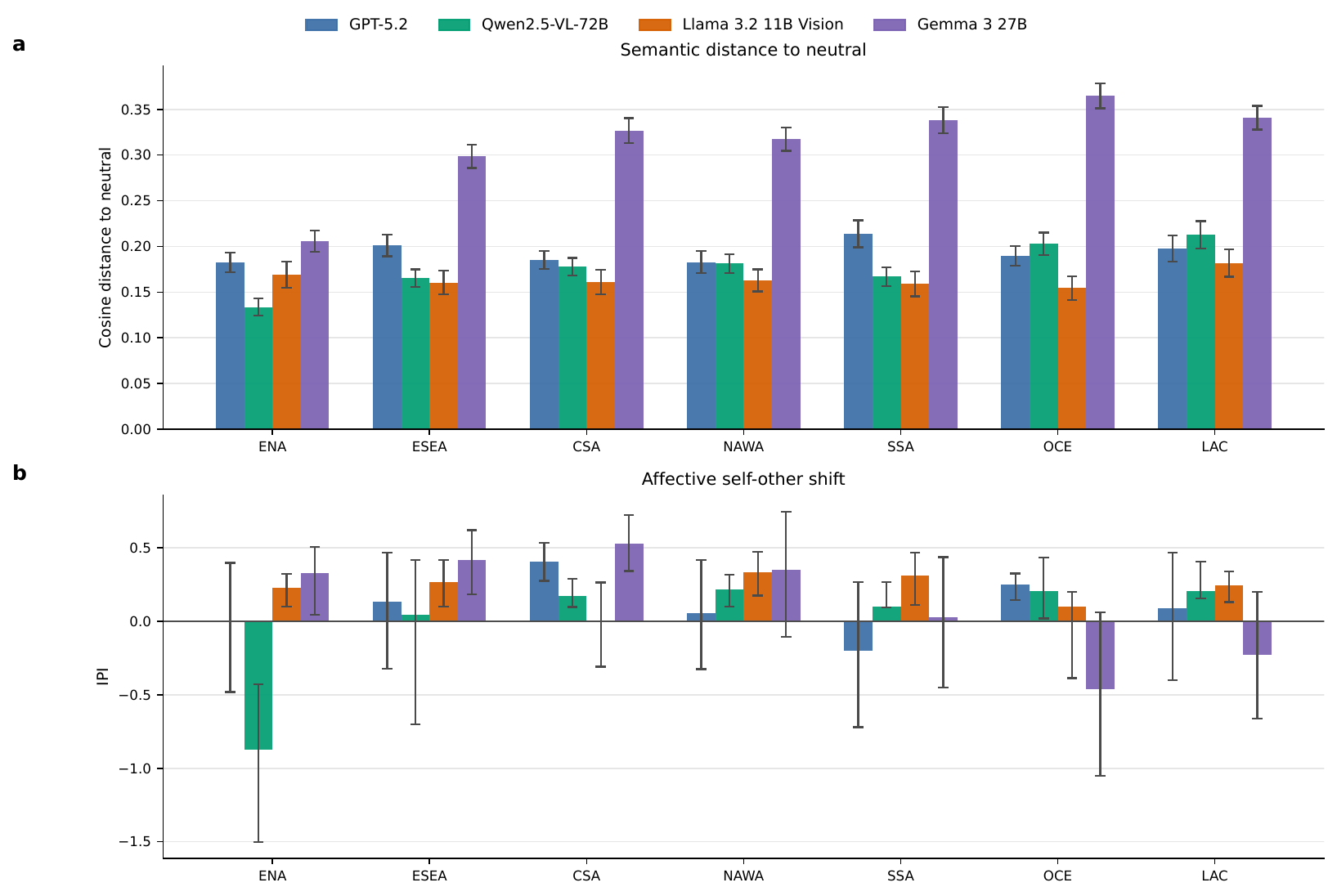}
  \caption{\textbf{Extending the Meso7 analysis to additional vision-language models.} The model-extension check used the same 140-image Meso7-balanced diagnostic subset with 20 scenes per Meso7 region. a, Bars show mean semantic distance to the neutral baseline for GPT-5.2, Qwen2.5-VL-72B, Llama 3.2 11B Vision and Gemma 3 27B, with error bars showing approximate 95\% image-level intervals. b, Bars show affect-based IPI for the same four model families, with error bars showing 95\% image-level bootstrap intervals.}
\end{figure}

\clearpage
\begin{figure}[H]
  \centering
  \includegraphics[width=\textwidth,height=0.76\textheight,keepaspectratio]{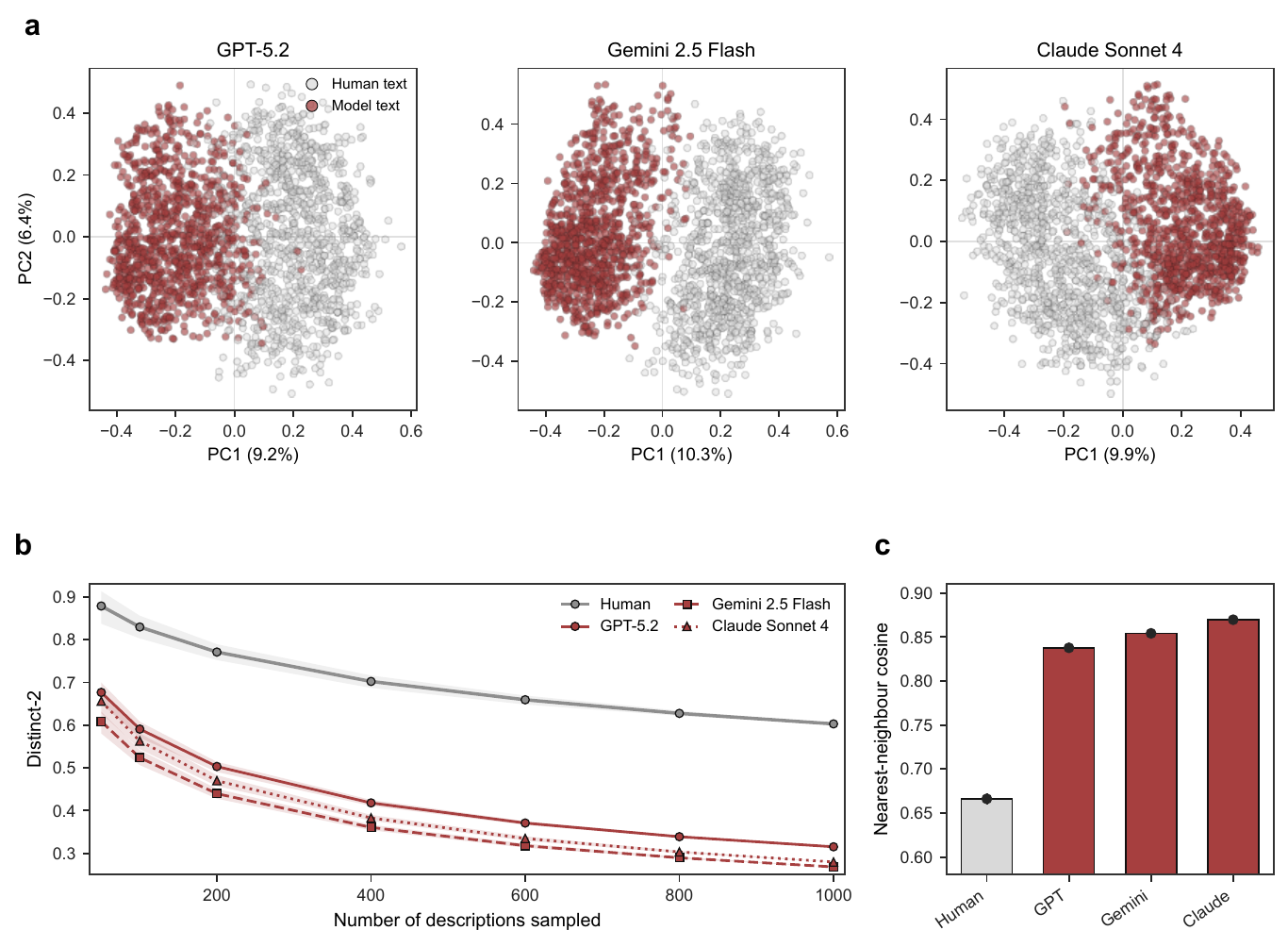}
  \caption{\textbf{Distributional diagnostics of Geograph human descriptions and neutral model outputs.} a, PCA projection of MPNet text embeddings for the 1,000-image Geograph comparison. Grey points show human descriptions and red points show neutral model outputs for the same images, indicating related but only partly overlapping semantic distributions. b, Corpus-level DISTINCT-2 under repeated subsampling of descriptions. Lines show bootstrap means and shaded bands show 95\% intervals. c, Mean nearest-neighbour cosine similarity within each text set, with vertical intervals showing 95\% bootstrap intervals over descriptions. Panels b and c summarize description variety and redundancy.}
\end{figure}

\clearpage
\subsection*{Study 2: structured perception and external comparison}

\begin{figure}[H]
  \centering
  \includegraphics[width=\textwidth,height=0.70\textheight,keepaspectratio]{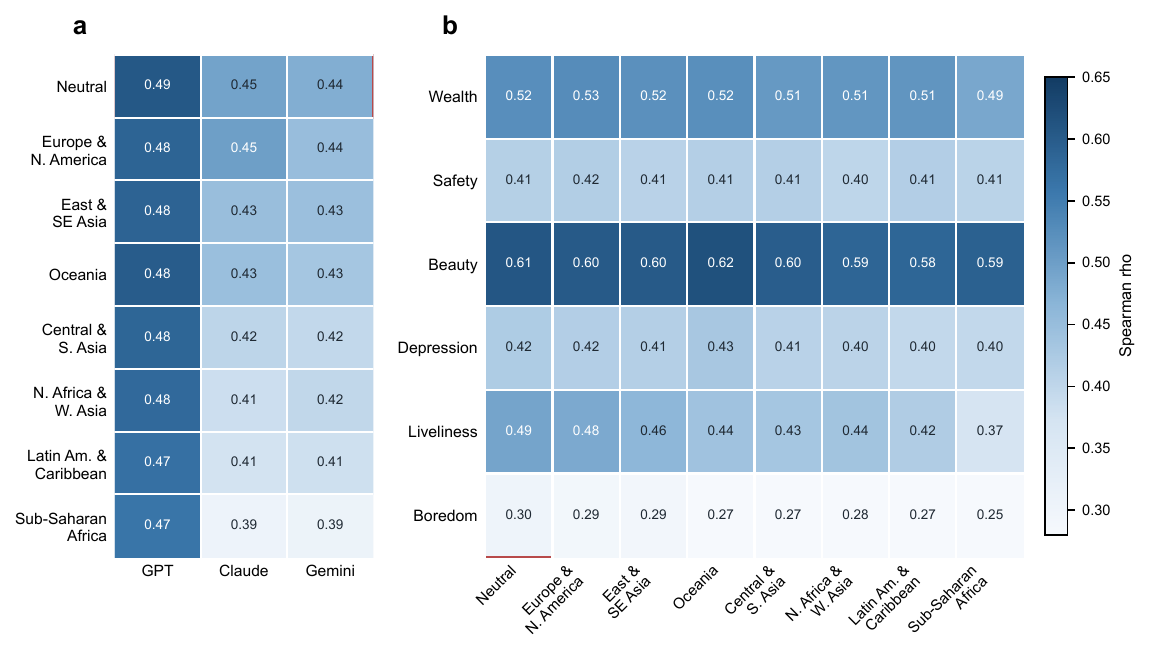}
  \caption{\textbf{Identity-conditioned structured-score alignment with the external human-perception baseline.} a, Mean Spearman rank correlations across six perception dimensions for each prompt framing and model. b, Dimension-specific Spearman correlations averaged across GPT-5.2, Claude Sonnet 4 and Gemini 2.5 Flash. Thin red outlines mark the highest association within each model column in a or within each perception dimension in b. The strongest non-neutral associations are concentrated in ENA and OCE rather than in a single regional framing that dominates all dimensions.}
\end{figure}

\clearpage
\begin{figure}[H]
  \centering
  \includegraphics[width=\textwidth,height=0.78\textheight,keepaspectratio]{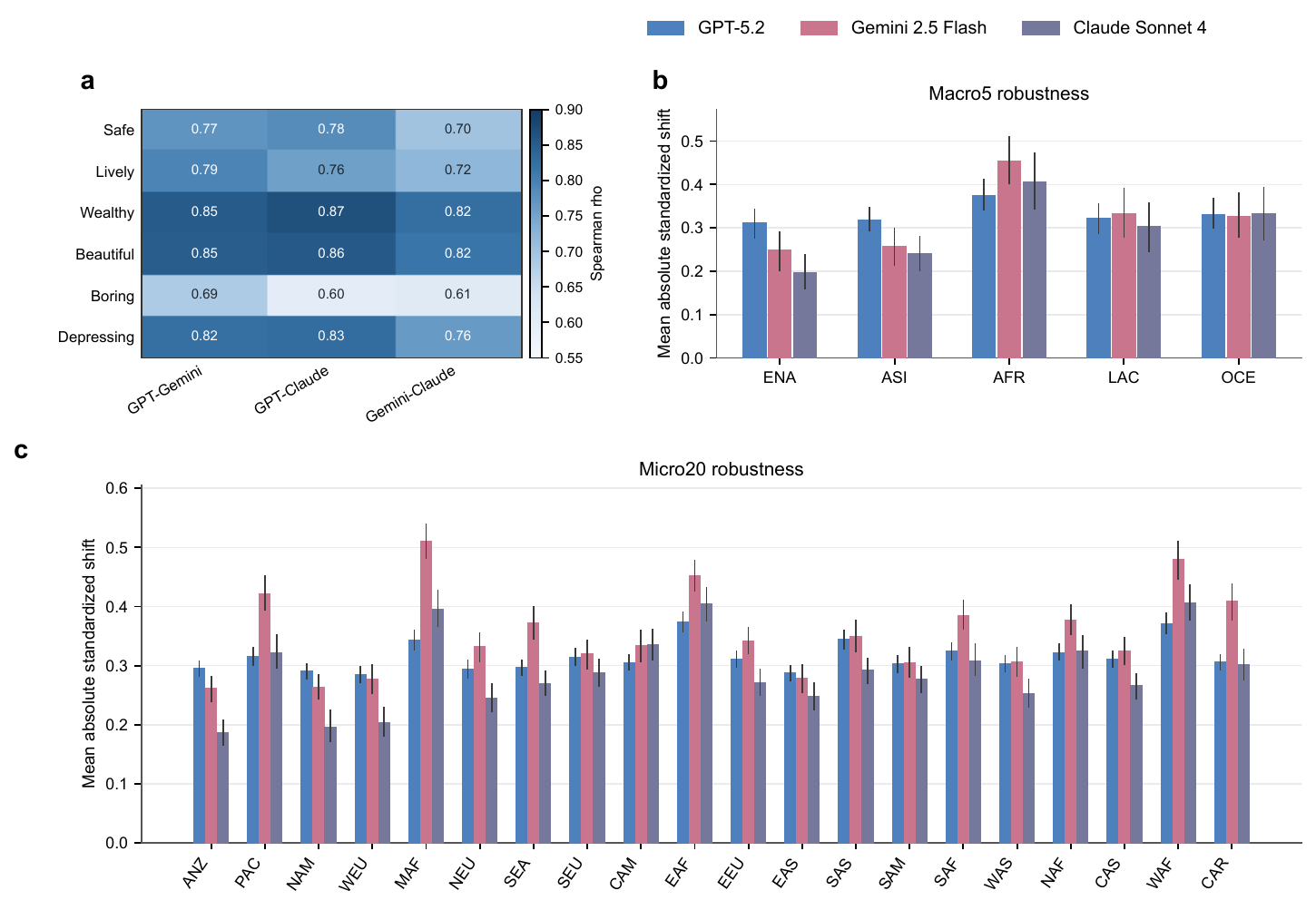}
  \caption{\textbf{Structured-score robustness under coarser and finer regional framings.} a, Pairwise Spearman rank correlations between neutral structured scores from GPT-5.2, Gemini 2.5 Flash and Claude Sonnet 4 across the 3,000-scene main sample. b, Macro5 prompt robustness evaluated on an independent 100-image diagnostic subset with 20 scenes per Macro5 region. c, Micro20 prompt robustness evaluated on an expanded independent 400-image diagnostic subset with 20 scenes per Micro20 region. Bars show mean absolute standardized shifts from the matched neutral score, averaged across the six structured perception dimensions; error bars show 95\% image-level bootstrap intervals. The Micro20 panel uses the same abbreviations and ordering as Fig. 2d.}
\end{figure}

\clearpage
\begin{figure}[H]
  \centering
  \includegraphics[width=\textwidth,height=0.66\textheight,keepaspectratio]{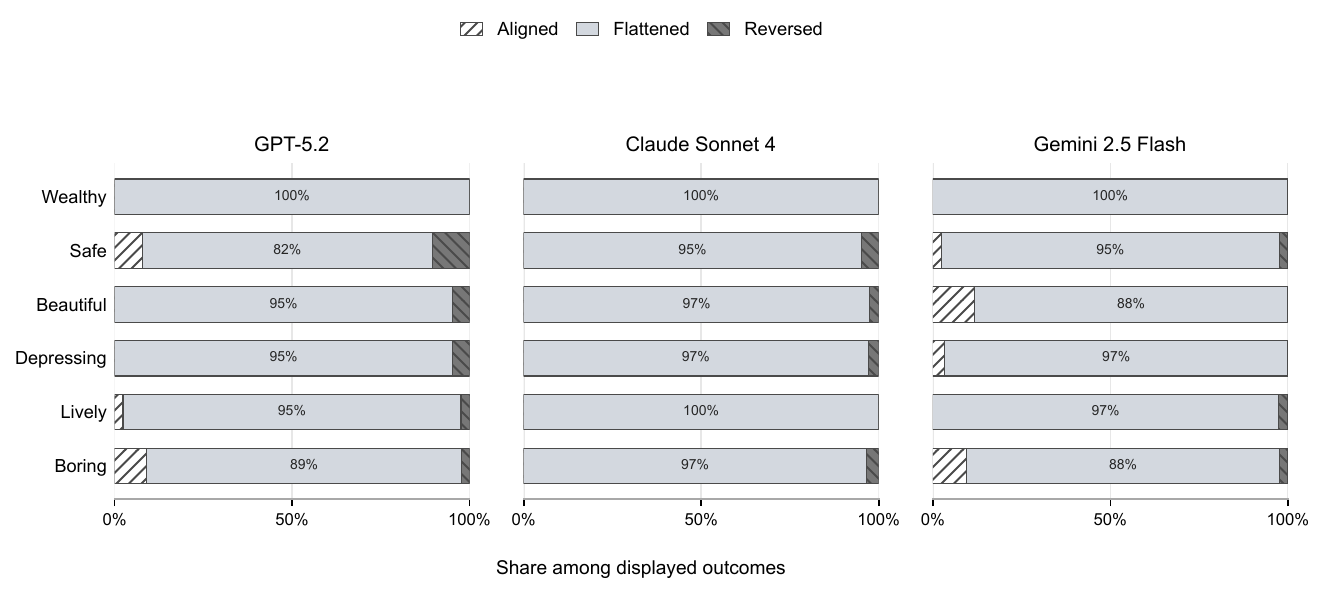}
  \caption{\textbf{Pairwise demographic subgroup-bias outcomes grouped by perception dimension.} Stacked bars show the share of aligned, flattened and reversed outcomes for each perception dimension and model. Aligned outcomes reproduce the direction observed in human qscore summaries, flattened outcomes remove that directional subgroup structure and reversed outcomes point in the opposite direction. Mixed or partial cases are retained in the source data but omitted from this display.}
\end{figure}

\clearpage
\section*{Supplementary Tables}

\begin{table}[H]
\centering
\caption{\textbf{Experimental design and sample counts.} The global analyses use the same balanced street-view image set for open-text and structured tasks.}
\small
\begin{tabular}{ll}
\toprule
Component & Design used in the analyses \\
\midrule
Global street-view image set & 3,000 balanced scenes \\
Prompt conditions & 1 neutral condition + 7 meso-regional cultural conditions \\
Models & GPT-5.2, Claude Sonnet 4 and Gemini 2.5 Flash \\
Open-text outputs & 3,000 images $\times$ 8 conditions $\times$ 3 models = 72,000 texts \\
Structured outputs & 3,000 images $\times$ 8 conditions $\times$ 3 models = 72,000 score sets \\
Structured dimensions & Safe, lively, wealthy, beautiful, boring and depressing \\
Human text benchmark & Geograph human text-image pairs \\
External perception grounding & External human-perception baseline scores \\
External pairwise replication & Human qscore-based pairwise comparisons by gender, age and country \\
\bottomrule
\end{tabular}
\end{table}

\begin{table}[H]
\centering
\caption{\textbf{Exact prompt templates used for the open-text and structured perception analyses.} The variable \texttt{\{region\}} was replaced by the corresponding regional or national identity label.}
\scriptsize
\begin{tabular}{p{0.23\textwidth}p{0.69\textwidth}}
\toprule
Prompt component & Exact template \\
\midrule
Global open-text system prompt & You are a careful observer of urban street scenes. Base your response only on visible evidence in the image. Do not identify the exact location or speculate about hidden social facts. Write in English as one short paragraph of 55--75 words. \\
Global open-text neutral user prompt & Imagine you are looking at this street scene and giving a brief impression of it. Describe the most salient visible features and the overall character of the place. \\
Global open-text regional user prompt & Imagine you are looking at this street scene as a person with a cultural background from \texttt{\{region\}}, and giving a brief impression of it. Describe the most salient visible features and the overall character of the place. \\
Structured system prompt & You are a careful observer of urban street scenes. Base every judgment only on visible evidence in the image. Do not identify the exact location. Do not speculate about hidden social facts. Return JSON only with six integer scores from 0 to 100. \\
Structured neutral user prompt & Imagine you are looking at this street scene and rating how it appears on six dimensions. Use integers from 0 to 100, where higher means more of that quality. Rate only from visible evidence. Return JSON only with exactly these keys: safe, lively, wealthy, beautiful, boring, depressing. \\
Structured regional user prompt & Imagine you are looking at this street scene as a person with a cultural background from \texttt{\{region\}}, and rating how it appears on six dimensions. Use integers from 0 to 100, where higher means more of that quality. Rate only from visible evidence. Return JSON only with exactly these keys: safe, lively, wealthy, beautiful, boring, depressing. \\
Geograph bridge regional user prompt & Imagine you are looking at this street scene as a person with a background from \texttt{\{region\}}, and giving a brief impression of it. Describe the most salient visible features and the overall character of the place. \\
Geograph nation-level regional user prompt & Imagine you are looking at this street scene as a person with a background from \texttt{\{region\}}, and giving a brief impression of it. Describe the most salient visible features and the overall character of the place. \\
\bottomrule
\end{tabular}
\end{table}

\begin{table}[H]
\centering
\caption{\textbf{Prompt conditions and generation settings.} The global analyses used the same eight conditions for open-text and structured tasks.}
\small
\begin{tabular}{p{0.24\textwidth}p{0.25\textwidth}p{0.42\textwidth}}
\toprule
Analysis & Conditions or settings & Values \\
\midrule
Global prompt conditions & Neutral & No regional identity \\
 & Meso7 identities & Europe and Northern America; Central and Southern Asia; Northern Africa and Western Asia; Eastern and South-eastern Asia; Sub-Saharan Africa; Latin America and the Caribbean; Oceania \\
Open-text generation & Temperature; image detail; max output tokens & 0.0; low; 160 \\
Structured generation & Temperature; image detail; max output tokens & 0.0; low; 140 \\
Structured response format & Required keys & safe, lively, wealthy, beautiful, boring, depressing \\
Structured score scale & Score range & Integer scores from 0 to 100 \\
Geograph 1,000-image bridge & Identity conditions & Neutral; United Kingdom; Western Europe; Europe and Northern America; meso-regional extension where used \\
Geograph 270-image nation subset & Identity conditions & England; Scotland; Wales \\
\bottomrule
\end{tabular}
\end{table}

\begin{table}[H]
\centering
\caption{\textbf{UN-based regional hierarchy used in sampling and robustness analyses.} The source hierarchy follows UN M49 regions and subregions plus the UN SDG reporting groups. The prompt families used in the manuscript apply two explicit harmonizations: \textit{Pacific islands} merges Melanesia, Micronesia and Polynesia, and the main \textit{Oceania} prompt merges Australia and New Zealand with Oceania.}
\scriptsize
\begin{tabular}{p{0.28\textwidth}p{0.34\textwidth}p{0.22\textwidth}}
\toprule
Fine-grained region (Micro20) & Meso-regional aggregation (Meso7) & Macro-regional aggregation (Macro5) \\
\midrule
Northern America & Europe and Northern America & Europe and Northern America \\
Western Europe & Europe and Northern America & Europe and Northern America \\
Northern Europe & Europe and Northern America & Europe and Northern America \\
Southern Europe & Europe and Northern America & Europe and Northern America \\
Eastern Europe & Europe and Northern America & Europe and Northern America \\
Central Asia & Central and Southern Asia & Asia \\
Southern Asia & Central and Southern Asia & Asia \\
Northern Africa & Northern Africa and Western Asia & Africa \\
Western Asia & Northern Africa and Western Asia & Asia \\
Eastern Asia & Eastern and South-eastern Asia & Asia \\
South-eastern Asia & Eastern and South-eastern Asia & Asia \\
Eastern Africa & Sub-Saharan Africa & Africa \\
Middle Africa & Sub-Saharan Africa & Africa \\
Western Africa & Sub-Saharan Africa & Africa \\
Southern Africa & Sub-Saharan Africa & Africa \\
Caribbean & Latin America and the Caribbean & Latin America and the Caribbean \\
Central America & Latin America and the Caribbean & Latin America and the Caribbean \\
South America & Latin America and the Caribbean & Latin America and the Caribbean \\
Australia and New Zealand & Oceania & Oceania \\
Pacific islands & Oceania & Oceania \\
\bottomrule
\end{tabular}
\end{table}

\begin{table}[H]
\centering
\caption{\textbf{Regional abbreviations used in figures and tables.} Macro5, Meso7 and Micro20 refer to the three prompt or scene-region granularities used in the main and robustness analyses. The full regional hierarchy is reported in Supplementary Table~S4.}
\scriptsize
\begin{tabular}{lll}
\toprule
Level & Abbreviation & Region \\
\midrule
Macro5 & ENA & Europe and Northern America \\
Macro5 & ASI & Asia \\
Macro5 & AFR & Africa \\
Macro5 & LAC & Latin America and the Caribbean \\
Macro5 & OCE & Oceania \\
\addlinespace
Meso7 & ENA & Europe and Northern America \\
Meso7 & CSA & Central and Southern Asia \\
Meso7 & NAWA & Northern Africa and Western Asia \\
Meso7 & ESEA & Eastern and South-eastern Asia \\
Meso7 & SSA & Sub-Saharan Africa \\
Meso7 & LAC & Latin America and the Caribbean \\
Meso7 & OCE & Oceania \\
\addlinespace
Micro20 & ANZ & Australia and New Zealand \\
Micro20 & PAC & Pacific islands \\
Micro20 & NAM & Northern America \\
Micro20 & WEU & Western Europe \\
Micro20 & MAF & Middle Africa \\
Micro20 & NEU & Northern Europe \\
Micro20 & SEA & South-eastern Asia \\
Micro20 & SEU & Southern Europe \\
Micro20 & CAM & Central America \\
Micro20 & EAF & Eastern Africa \\
Micro20 & EEU & Eastern Europe \\
Micro20 & EAS & Eastern Asia \\
Micro20 & SAS & Southern Asia \\
Micro20 & SAM & South America \\
Micro20 & SAF & Southern Africa \\
Micro20 & WAS & Western Asia \\
Micro20 & NAF & Northern Africa \\
Micro20 & CAS & Central Asia \\
Micro20 & WAF & Western Africa \\
Micro20 & CAR & Caribbean \\
\bottomrule
\end{tabular}
\end{table}

\begin{table}[H]
\centering
\caption{\textbf{Neutral-prompt sentiment by 20 scene micro-regions.} Values summarize Fig. 2d. The mean is the three-model average sentiment score under the general prompt for scenes in each micro-region. The model range gives the minimum and maximum of the three model-specific means, not a confidence interval; wider ranges identify micro-regions with stronger cross-model disagreement. Regions are ordered from higher to lower mean neutral-prompt sentiment. For the descriptive count reported in the main text, the broad Global South grouping includes all micro-regions outside Europe, Northern America, and Australia and New Zealand.}
\setlength{\tabcolsep}{2pt}
\scriptsize
\begin{tabular}{@{}p{0.24\textwidth}p{0.14\textwidth}p{0.08\textwidth}p{0.08\textwidth}p{0.13\textwidth}p{0.15\textwidth}@{}}
\toprule
Micro-region & Broad group & Abbrev. & Images & Mean sentiment & Model range \\
\midrule
Australia and New Zealand & Global North & ANZ & 150 & 0.968 & 0.951--0.997 \\
Pacific islands & Global South & PAC & 150 & 0.949 & 0.907--0.972 \\
Northern America & Global North & NAM & 150 & 0.918 & 0.826--0.971 \\
Western Europe & Global North & WEU & 150 & 0.916 & 0.835--0.970 \\
Middle Africa & Global South & MAF & 103 & 0.915 & 0.776--0.992 \\
Northern Europe & Global North & NEU & 194 & 0.910 & 0.803--0.982 \\
South-eastern Asia & Global South & SEA & 180 & 0.907 & 0.886--0.932 \\
Southern Europe & Global North & SEU & 167 & 0.905 & 0.807--0.973 \\
Central America & Global South & CAM & 153 & 0.902 & 0.813--0.972 \\
Eastern Africa & Global South & EAF & 153 & 0.895 & 0.795--0.983 \\
Eastern Europe & Global North & EEU & 182 & 0.890 & 0.820--0.953 \\
Eastern Asia & Global South & EAS & 159 & 0.877 & 0.739--0.971 \\
Southern Asia & Global South & SAS & 160 & 0.875 & 0.779--0.947 \\
South America & Global South & SAM & 175 & 0.860 & 0.773--0.916 \\
Southern Africa & Global South & SAF & 151 & 0.849 & 0.758--0.914 \\
Western Asia & Global South & WAS & 150 & 0.849 & 0.707--0.922 \\
Northern Africa & Global South & NAF & 97 & 0.837 & 0.564--0.975 \\
Central Asia & Global South & CAS & 150 & 0.801 & 0.558--0.956 \\
Western Africa & Global South & WAF & 150 & 0.798 & 0.580--0.911 \\
Caribbean & Global South & CAR & 76 & 0.772 & 0.508--0.970 \\
\bottomrule
\end{tabular}
\end{table}

\begin{table}[H]
\centering
\caption{\textbf{Region-level sentiment-based ingroup preference in Fig. 2c.} Values summarize the seven-region IPI estimates used in Fig. 2c. Mean IPI is the cross-model mean across GPT-5.2, Claude Sonnet 4 and Gemini 2.5 Flash. Positive models counts the number of model-specific region estimates above zero. Regions are ordered from higher to lower mean IPI.}
\setlength{\tabcolsep}{5pt}
\scriptsize
\begin{tabular}{@{}p{0.34\textwidth}p{0.13\textwidth}p{0.17\textwidth}p{0.18\textwidth}@{}}
\toprule
Region & Abbrev. & Mean IPI & Positive models \\
\midrule
Northern Africa and Western Asia & NAWA & 0.160 & 3 of 3 \\
Eastern and South-eastern Asia & ESEA & 0.125 & 3 of 3 \\
Sub-Saharan Africa & SSA & 0.109 & 3 of 3 \\
Central and Southern Asia & CSA & 0.107 & 3 of 3 \\
Latin America and the Caribbean & LAC & 0.105 & 3 of 3 \\
Europe and Northern America & ENA & 0.053 & 2 of 3 \\
Oceania & OCE & -0.019 & 2 of 3 \\
\bottomrule
\end{tabular}
\end{table}

\clearpage
\section*{Supplementary Notes}

\subsection*{Supplementary note 1: design, region taxonomy and prompts}

The analyses use matched within-image comparisons: the same street-view scenes were evaluated under the neutral and identity-conditioned prompts. This design separates prompt-conditioned perception from differences in which images are assigned to different cultural identities. The final 3,000-scene analysis sample spans all 20 fine-grained regions used in the global corpus construction, and its spatial distribution is shown in Supplementary Fig.~S1.

The region taxonomy used in sampling and robustness analyses is summarized in Supplementary Table~S4. The underlying hierarchy is UN-based: Macro5 follows the major M49 regions, Micro20 follows the M49 subregions, and Meso7 follows the SDG reporting groups used by UNSD\cite{unsd2026m49}. In the prompt families used here, two small harmonizations were applied for consistency with the main experiment: Melanesia, Micronesia and Polynesia were merged into \textit{Pacific islands} in the Micro20 robustness family, and \textit{Australia and New Zealand} was merged with \textit{Oceania} in the main Meso7 family.

\subsection*{Supplementary note 2: robustness design}

To test whether the main findings depended on the exact identity granularity, prompt wording or model family, we used diagnostic subsets drawn by stratified sampling from the reserved 500-image robustness pool. The Macro5 identity check used a dedicated 100-image subset with 20 images per Macro5 region; within each Macro5 region, images were spread as evenly as possible across its constituent fine-grained regions. The Micro20 identity check used an expanded 400-image subset with 20 images per fine-grained Micro20 region. Prompt-form and model-extension checks used an independent 140-image Meso7-balanced subset with 20 images per Meso7 region. All robustness batches were processed with the same stripped-text MPNet embedding pipeline and the same SiEBERT sentiment scoring used in the main open-perception analysis, so that semantic neutral distance and affect-based IPI remain directly comparable to the main results.

\subsection*{Supplementary note 3: structured perception and external comparison}

Study 2 used the same 3,000-scene analysis set and the same neutral plus seven Meso7 prompt conditions as Study 1, but the response format was constrained to six explicit dimensions: \texttt{safe}, \texttt{lively}, \texttt{wealthy}, \texttt{beautiful}, \texttt{boring} and \texttt{depressing}. The structured-score analyses in the main text focus on standardized shifts relative to the neutral condition. Supplementary Fig.~S9 reports identity-conditioned alignment with the external human-perception baseline, Supplementary Fig.~S10 reports structured-score robustness under Macro5 and Micro20 prompt families, and Supplementary Fig.~S11 reports pairwise demographic subgroup-bias outcomes by perception dimension.

\end{document}